\documentclass[letterpaper,10pt,conference]{ieeeconf}

\IEEEoverridecommandlockouts

\newif\ifarxivmode
\arxivmodetrue

\newif\ifanonymous
\anonymousfalse

\ifarxivmode
\else
  \usepackage{titlesec}
  \titlespacing*{\section}{0pt}{8pt plus 2pt minus 2pt}{4pt plus 2pt minus 2pt}
  \titlespacing*{\subsection}{0pt}{6pt plus 2pt minus 2pt}{3pt plus 2pt minus 2pt}

  \let\oldthebibliography\thebibliography
  
  \renewcommand{\thebibliography}[1]{%
    \oldthebibliography{#1}%
    \setlength{\itemsep}{0pt}%
    \setlength{\parskip}{0pt}%
  }

\fi

\PassOptionsToPackage{table}{xcolor}
\usepackage{amsmath}
\usepackage{amssymb}
\usepackage{amsfonts}
\usepackage{mathtools, cuted} %
\usepackage{bm}
\usepackage{empheq}
\usepackage{calc}
\usepackage{bbm}

\usepackage[usenames,dvipsnames]{xcolor}
\usepackage{graphicx}
\usepackage{svg}
\usepackage{caption}
\usepackage{subcaption}

\usepackage{tabularx}
\usepackage{multirow}
\usepackage{makecell}
\usepackage{booktabs}
\usepackage{hhline}
\usepackage{arydshln}
\usepackage{longtable}

\newcolumntype{H}{>{\setbox0=\hbox\bgroup}c<{\egroup}@{}}

\usepackage{algorithm}
\usepackage{algpseudocode}

\usepackage[sort,compress]{cite}
\usepackage[pdfstartview={FitV},pdfpagelayout={TwoColumnLeft},bookmarksopen=true,plainpages=false,colorlinks=true,linkcolor=black,citecolor=blue,urlcolor=blue,filecolor=black,pagebackref=false,hypertexnames=false,pdfpagelabels]{hyperref}
\usepackage[T1]{fontenc}
\usepackage[nameinlink,capitalize]{cleveref}
\usepackage[nolist]{acronym}

\usepackage[]{siunitx}
\usepackage{esvect}
\usepackage{xspace}
\usepackage{enumerate}
\usepackage{balance}
\usepackage{etoolbox}
\usepackage{textcomp}
\usepackage{mdframed}
\usepackage{footnote}
\makesavenoteenv{tabular}
\makesavenoteenv{table}

\robustify\bfseries

\ifdefined\qtyproduct
\else
  \ifdefined\NewCommandCopy
    \NewCommandCopy\qtyproduct\SI
  \else
    \NewDocumentCommand\qtyproduct{O{}mm}{\SI[#1]{#2}{#3}}
  \fi
\fi

\usepackage{tikz}
\usetikzlibrary{shapes,snakes}
\usetikzlibrary{tikzmark}
\usetikzlibrary{calc, arrows.meta, positioning}
\usetikzlibrary{shapes.multipart}
\usetikzlibrary{patterns}

\usepackage{setspace}
\ifarxivmode
\else
\fi

\DeclareCaptionLabelSeparator{periodspace}{.\quad}
\newcommand{\ie}{i.e.,\xspace}
\newcommand{\eg}{e.g.,\xspace}

\newcommand{\best}[1]{\textbf{\textcolor{ForestGreen}{#1}}}

\newcommand{\worst}[1]{\textbf{\textcolor{red}{#1}}}

\newcommand{\rshade}[2][red!12]{\colorbox{#1}{\strut #2}}

\definecolor{opt_color}{RGB}{203,255,182}
\definecolor{check_color}{RGB}{195,218,255}
\definecolor{recheck_color}{RGB}{255,176,176}
\definecolor{delaycheck_color}{RGB}{255,228,181}

\definecolor{agent1_color}{RGB}{234,153,153}
\definecolor{agent2_color}{RGB}{151,104,175}
\definecolor{agent3_color}{RGB}{249,203,156}
\definecolor{agent4_color}{RGB}{194,115,156}
\definecolor{agent5_color}{RGB}{159,197,232}
\definecolor{agent6_color}{RGB}{117,186,117}
\definecolor{obstacle_color}{RGB}{160,117,109}

\newif\ifchangehighlightedblue
\changehighlightedbluefalse

\newcommand{\changetext}[1]{%
  \ifchangehighlightedblue%
    \textcolor{blue}{#1}%
  \else%
    #1%
  \fi%
}

\newenvironment{changeframe}{%
  \ifchangehighlightedblue%
    \begin{mdframed}[linecolor=blue, linewidth=0.8pt, backgroundcolor=white, innerleftmargin=6pt, innerrightmargin=6pt, innertopmargin=6pt, innerbottommargin=6pt]%
  \else%
    \begingroup%
  \fi%
}{%
  \ifchangehighlightedblue%
    \end{mdframed}%
  \else%
    \endgroup%
  \fi%
}

\newacro{slam}[SLAM]{Simultaneous Localization and Mapping}
\newacro{uav}[UAV]{Unmanned Aerial Vehicle}
\acrodefplural{uav}[UAVs]{unmanned aerial vehicles}
\newacro{gns}[GNS]{Global Navigation Satellite}
\newacro{gnss}[GNSS]{Global Navigation Satellite System}
\acrodefplural{gnss}[GNSS]{Global Navigation Satellite Systems}
\newacro{mcl}[MCL]{Monte-Carlo localization}
\newacro{imu}[IMU]{Inertial Measurement Unit}
\newacro{dof}[DOF]{degree-of-freedom}
\acrodefplural{dof}[DOFs]{degrees-of-freedom}
\newacro{ransac}[RANSAC]{random sample consensus}
\newacro{map}[MAP]{maximum a posteriori}
\newacro{mle}[MLE]{maximum likelihood estimation}
\newacro{rms}[RMS]{root-mean-square}
\newacro{dem}[DEM]{digital elevation model}
\newacro{vio}[VIO]{visual-inertial odometry}
\newacro{cnn}[CNN]{convolutional neural network}
\newacro{pdf}[pdf]{probability density function}
\newacro{ahrs}[AHRS]{attitude and heading reference system}
\newacro{lidar}[LIDAR]{light detection and ranging}
\newacro{relu}[ReLU]{rectified linear unit}
\newacro{rtk}[RTK]{real-time kinematic}
\newacro{gps}[GPS]{global positioning system}
\newacro{fcn}[FCN]{fully-connected network}
\newacro{brm}[BRM]{building ratio map}
\newacro{sfm}[SfM]{Structure-from-Motion}
\newacro{vpr}[VPR]{visual place recognition}
\newacro{fov}[FOV]{field of view}
\newacro{poc}[POC]{partially overlapping circular}

\title{\LARGE \bf MIGHTY: Hermite Spline-based Efficient Trajectory Planning}

\ifanonymous
  \author{
  Anonymous Submission
  }
\else
  \author{
  Kota Kondo$^{1}$, Yuwei Wu$^{2}$, Vijay Kumar$^{2}$, Jonathan P.\ How$^{1}$%
  \thanks{$^{1}$Department of Aeronautics and Astronautics, Massachusetts Institute of Technology (\texttt{\{kkondo,jhow\}@mit.edu}).
  $^{2}$Department of Electrical and Systems Engineering, University of Pennsylvania (\texttt{\{yuweiwu,kumar\}@seas.upenn.edu}).}
  }
\fi

\begin{document}

\ifarxivmode
\pagenumbering{arabic}
\fi
\maketitle
\vspace{-4em}
\thispagestyle{plain}
\pagestyle{plain}

\begin{abstract}
\acresetall
Hard-constraint trajectory planners often rely on commercial solvers and demand substantial computational resources.
Existing soft-constraint methods achieve faster computation, but either (1) decouple spatial and temporal optimization or (2) restrict the search space.
To overcome these limitations, we introduce \textbf{MIGHTY}, a Hermite spline-based planner that performs spatiotemporal optimization while fully leveraging the continuous search space of a spline.
In simulation, MIGHTY achieves a 9.3\% reduction in computation time and a 13.1\% reduction in travel time over state-of-the-art baselines, with a 100\% success rate.
In hardware, MIGHTY completes multiple high-speed flights up to 6.7 m/s in a cluttered static environment and long-duration flights with dynamically added obstacles.

\end{abstract}

\ifanonymous
\else
  \section*{Supplementary Material}
  \noindent\textbf{Video}: \href{https://youtu.be/Pvb-VPUdLvg}{https://youtu.be/Pvb-VPUdLvg} \\
  \noindent\textbf{Code}: \href{https://github.com/mit-acl/mighty.git}{https://github.com/mit-acl/mighty.git}
\fi

\section{Introduction}\label{sec:introduction}

Trajectory planning for autonomous navigation has been extensively studied with various parameterizations~\cite{mellinger2011minimum,7138978,liu2017planning,9147300,kondo2025dynus,zhou2019robust,zhou2021ego-planner,zhou2022swarm,zhou2021raptor,tordesillas2022faster,richter2016polynomial,levy2024stitcher,wang2022geometrically,ren2025super}.
Hard-constraint approaches~\cite{mellinger2011minimum,7138978,liu2017planning,9147300,kondo2025dynus} explicitly enforce safety but require computationally intensive solvers unsuitable for high-frequency replanning.
Soft-constraint planners (EGO-Planner~\cite{zhou2021ego-planner}, RAPTOR~\cite{zhou2021raptor}, SUPER~\cite{ren2025super}) achieve faster convergence.
Some jointly optimize geometry and timing~\cite{wang2022geometrically,ren2025super}, while others decouple path and time allocation~\cite{richter2016polynomial,zhou2021raptor}.
Increasing decision variables improves performance but enlarges the problem.
Building on these ideas, we introduce \textbf{MIGHTY}, a Hermite spline-based planner.

\begin{table*}[!t]
  \renewcommand{\arraystretch}{0.9}
  \scriptsize
  \begin{centering}
  \caption{\centering State-of-the-art Unconstrained UAV Trajectory Planners}
  \label{tab:state_of_the_art_comparison}
  \begin{changeframe}
  \resizebox{\textwidth}{!}{
  \begin{tabular}{>{\centering\arraybackslash}m{0.15\textwidth}
                  >{\centering\arraybackslash}m{0.12\textwidth}
                  >{\centering\arraybackslash}m{0.14\textwidth}
                  >{\centering\arraybackslash}m{0.07\textwidth}
                  >{\centering\arraybackslash}m{0.20\textwidth}
                  >{\centering\arraybackslash}m{0.05\textwidth}
                  >{\centering\arraybackslash}m{0.10\textwidth}}
  \toprule
  \multirow{2}{*}[-1.1ex]{\textbf{Method}} & \multirow{2}{*}[-1.1ex]{\textbf{Representation}} & \multirow{2}{*}[-1.1ex]{\textbf{Variables}} & \multirow{2}{*}[-1.1ex]{\textbf{Time Opt.}} & \multirow{2}{*}[-1.1ex]{\textbf{Search Space}} & \multicolumn{2}{c}{\textbf{Local Control}} \tabularnewline
  \cmidrule(lr){6-7}
   & & & & & \textbf{Geom.} & \textbf{Deriv.} \tabularnewline
  \midrule
  \textbf{Polynomial} \cite{richter2016polynomial} (2016)
    & 9th-order polynomial
    & endpoint derivs, time
    & \textcolor{red}{decoupled}
    & \textcolor{red}{limited polynomial+time space}
    & \textcolor{red}{no (global)}
    & \textcolor{red}{direct but global} \tabularnewline
  \midrule
  \textbf{Robust} \cite{zhou2019robust} (2019)
    & non-uniform B-spline
    & B-spline cntrl pts, time
    & \textcolor{red}{decoupled}
    & \textcolor{red}{limited polynomial+time space}
    & \textcolor{ForestGreen}{yes}
    & \textcolor{red}{indirect} \tabularnewline
  \midrule
  \textbf{EGO-Planner} \cite{zhou2021ego-planner} (2021)
    & uniform B-spline
    & B-spline cntrl pts
    & \textcolor{red}{fixed}
    & \textcolor{red}{limited polynomial+time space}
    & \textcolor{ForestGreen}{yes}
    & \textcolor{red}{indirect} \tabularnewline
  \midrule
  \textbf{RAPTOR} \cite{zhou2021raptor} (2021)
    & uniform B-spline
    & B-spline cntrl pts
    & \textcolor{red}{fixed}
    & \textcolor{red}{limited polynomial+time space}
    & \textcolor{ForestGreen}{yes}
    & \textcolor{red}{indirect} \tabularnewline
  \midrule
    \textbf{MINCO class} \cite{wang2022geometrically, zhou2022swarm, ren2025super} (2022, 2022, 2025)
    & MINCO
    & pos., time
    & \textcolor{ForestGreen}{joint}
    & \makecell{\textcolor{red}{limited polynomial+time space} \\ \textcolor{red}{(MINCO class)}}
    & \textcolor{red}{no (global)}
    & \textcolor{red}{indirect} \tabularnewline
  \midrule
  \textbf{MIGHTY (proposed)}
    & Hermite spline
    & Hermite cntrl pts, time
    & \textcolor{ForestGreen}{joint}
    & \textcolor{ForestGreen}{full polynomial+time space}
    & \textcolor{ForestGreen}{yes}
    & \textcolor{ForestGreen}{direct \& local} \tabularnewline
  \bottomrule
  \end{tabular}
  }
  \end{changeframe}
  \par\end{centering}
  \vspace{-2.3em}
\end{table*}

\subsection{Review of Unconstrained Frameworks}\label{subsec:related_work}

Differential flatness~\cite{FLIESS01061995} enables efficient optimization of quadrotor trajectories by parameterizing the flat output and their derivatives.
Given fixed timestamps at trajectory waypoints, minimum control effort (\eg jerk/snap) problems using piecewise polynomial representation can be formulated as quadratic programs (QPs), which are convex and can be solved efficiently~\cite{mellinger2011minimum}.
However, when segment durations are included as optimization variables (time allocation), the resulting joint problem becomes substantially more complex, yielding a nonlinear program (NLP) that is often ill-conditioned because of the coupling between spatial and temporal variables.

Decoupling into two stages enables unconstrained formulations by replacing hard constraints with soft penalties.
Richter et al.~\cite{richter2016polynomial} use 9\textsuperscript{th}-order polynomials with endpoint derivatives: a QP minimizes snap, then gradient-based time allocation adjusts duration.
\changetext{Endpoint derivatives are decision variables, so there is direct derivative parameterization.
However, the free derivatives are globally coupled, so there is no local control.}
Collision avoidance inserts mid-segment waypoints, triggering repeated solves.
Extensions~\cite{7353622,7759784} improve stability and enable real-time replanning via ESDF-based potential costs.

Another standard unconstrained framework uses B-splines for trajectory representation, with time-related variables encoded in a knot vector.
\changetext{B-splines offer inherent smoothness and geometric local control: changing one control point affects only a bounded neighborhood of the curve.
The convex hull property enables efficient collision checking and enforces dynamic feasibility~\cite{9147300, zhou2019robust, zhou2021ego-planner, zhou2021raptor}.
However, B-splines do not provide direct local control of higher-order derivatives.
Derivatives at any curve point are implicit functions of multiple neighboring control points.
One can bound derivatives efficiently via the convex hull property, but cannot independently prescribe a specific velocity or acceleration at a given knot through a single variable.
This geometric local control property has been widely exploited for efficient trajectory refinement and replanning in cluttered environments~\cite{jung2013line, pan2012collision, usenko2017real, ding2018trajectory}.}
Zhou et al.~\cite{zhou2019robust} parameterize trajectories as non-uniform cubic B-splines initialized from kinodynamic A* paths.
The nonlinear optimization minimizes the cost considering smoothness, a repulsive collision potential from a Euclidean distance field (EDF), and soft penalties on derivative control points exceeding maximum velocity and acceleration.
Finally, an iterative time-adjustment method rescales knot spans of the non-uniform spline to guarantee dynamic feasibility.
EGO-Planner~\cite{zhou2021ego-planner} and RAPTOR~\cite{zhou2021raptor} simplify the representation by using uniform B-splines with fixed knot spans, leveraging convex hull properties to enforce velocity, acceleration, and jerk constraints directly on control points.
EGO-Planner optimizes trajectories via a quasi-Newton solver (L-BFGS)~\cite{liu1989lbfgs} for fast gradient-based convergence, with an additional time-reallocation and anisotropic curve-fitting refinement stage that resizes knot spans and fits a new spline to ensure feasibility and smoothness.
RAPTOR also employs a two-stage process: a closed-form QP warm-start for initial smoothness, followed by a nonlinear refinement phase that penalizes collision via an ESDF and ensures dynamic limits via control point constraints.

\changetext{%
Wang et al.~\cite{wang2021glst} employ diffeomorphism between coefficients and derivatives with linear-complexity gradients.
MINCO~\cite{wang2022geometrically} builds on this: coefficients are analytically determined by waypoints and durations, jointly optimized via linear-complexity solver for minimum jerk/snap.
MINCO excels when minimum control effort dominates, but searches only the MINCO class.
MINCO's waypoint positions are decision variables, but polynomial coefficients are determined through a banded linear system that couples adjacent segments, so there is no geometric local control.
Higher-order derivatives are not decision variables, and they are analytically determined from waypoints and durations, so there is no direct higher-derivative control.
Under added objectives, this restriction can yield suboptimal performance compared to more general representations.
}
EGO-Swarm2~\cite{zhou2022swarm} and SUPER~\cite{ren2025super} adopt MINCO; SUPER adds exploratory/backup trajectories for safety.

\changetext{%
MIGHTY optimizes positions and derivatives at each knot and time durations as explicit variables, enabling local control of waypoints and higher-order dynamics.
}

\subsection{MIGHTY Contributions}\label{subsec:contributions}

Table~\ref{tab:state_of_the_art_comparison} summarizes the discussion above. Unconstrained higher-order polynomial representations~\cite{richter2016polynomial} are typically too slow for real-time planning.
Uniform or non-uniform splines~\cite{zhou2019robust, zhou2021ego-planner, zhou2021raptor} can generate trajectories faster compared to spatiotemporal joint optimization; however, they can generate sub-optimal trajectories due to decoupled time adjustment.
\changetext{Waypoint-and-duration-based parameterizations~\cite{wang2022geometrically, zhou2022swarm, ren2025super} are restricted to the low-dimensional subspace induced by the MINCO parameterization, lack direct parameterization of higher-order derivatives at individual knots, and exhibit waypoint propagation where changes to one waypoint affect multiple segments through the globally coupled coefficient system.
While these effects enable compact representation and analytical optimality for minimum-control objectives, they can complicate constraint/cost handling beyond minimum-control and limit local control of trajectory shape and higher-order dynamics.}

To address the limitations of existing soft-constraint planners, \textbf{MIGHTY} uses a Hermite-spline representation and an unconstrained nonlinear optimizer that searches directly over spatial waypoints, endpoint derivatives, and segment durations.
\changetext{A Hermite spline representation guarantees continuity without global coefficient coupling, which fundamentally enables direct local control over both positions and higher-order dynamics at each knot. These properties could be exploited for efficient trajectory refinement (as in B-spline methods~\cite{jung2013line, pan2012collision, usenko2017real, ding2018trajectory}) but are not available in MINCO's globally coupled parameterization.}
The key contributions of the paper are:
\begin{enumerate}
    \item \textbf{Hermite Spline Planner:}
    \changetext{Joint spatiotemporal optimization in a single solve, with explicit parameterization of positions and derivatives at each knot.
    This representation enables local control of higher-order dynamics and trajectory shape, which the optimizer can leverage when needed.
    The approach, \textbf{MIGHTY}, achieves fast solve times and lower travel time than baselines.}
    \item \textbf{Comprehensive Simulation Study:}
    We benchmark in both simple and complex scenes, comparing against state-of-the-art methods in static environments and validating safe behavior in dynamic environments. MIGHTY achieves \best{\SI{9.3}{\%}} reduction in computation time and \best{\SI{13.1}{\%}} reduction in travel time in trajectory representation benchmarking, and it achieves the shortest travel time and path length while maintaining the \best{\SI{100}{\%}} success rate.
    \item \textbf{Extensive Hardware Experiments:}
    Long-duration and high-speed flights, and dynamic obstacle avoidance scenarios using a LiDAR-based perception and localization system.
    MIGHTY achieves collision-free flights and a top speed of \best{\SI{6.7}{\m/\s}} in high-speed flight experiments.
\end{enumerate}

\section{Hermite Spline}\label{sec:hermite_spline}

This section introduces the Hermite-spline notation used in MIGHTY's formulation.
MIGHTY uses a degree-$d$ Hermite spline with $M$ segments.
For odd degree $d=2\nu+1$, each segment is parameterized by the knot values of the position and its derivatives up to order $\nu$.
For example, a quintic Hermite spline ($d=5$, $\nu=2$) specifies position, velocity, and acceleration.
Across knots, a Hermite spline automatically guarantees $\mathcal C^\nu$ continuity by sharing the knot values.
Although Hermite splines apply in any dimension, we work in $\mathbb{R}^3$ for quadrotor planning and use $d=5$ in this paper.
For each segment $s = 0, \dots, M-1$, we define a normalized time variable $\tau_s \in [0,1]$ on segment $s$ as
$
\tau_s=(t - t_s)/T_s\, \in[0,1].
$
Then each segment can be denoted as:
\(
\mathbf{x}(t) = \mathbf{x}_s(\tau_s), \, t \in \bigl[t_s, \; t_{s+1} \bigr],
\)
where $t_s := \sum_{r=0}^{s-1} T_r$, and $T_s$ is the duration of segment $s$.
Then we write
\begin{equation}
\label{eq:hermite_segment}
\quad  \mathbf{x}_s(\tau_s) = \sum_{k=0}^{5} h_k(\tau_s) \, \mathbf{H}_{s,k},
\end{equation}
where
\(
\mathbf{H}_{s,0} = \mathbf{p}_s, 
\mathbf{H}_{s,1} = T_s \, \mathbf{v}_s,
\mathbf{H}_{s,2} = \tfrac{1}{2} T_s^2 \, \mathbf{a}_s,
\mathbf{H}_{s,3} = \mathbf{p}_{s+1}, \
\mathbf{H}_{s,4} = T_s \, \mathbf{v}_{s+1}, \
\mathbf{H}_{s,5} = \tfrac{1}{2} T_s^2 \, \mathbf{a}_{s+1}.
\)
Here, \(h_k(\tau_s)\) are the standard quintic Hermite basis functions.
The \(\mathbf{p}_i,\mathbf{v}_i,\mathbf{a}_i\in\mathbb{R}^3 \; (i=0,\dots,M) \) and \(
T_s>0 \;(s=0,\dots,M-1)\) are the interior knot positions, velocities, accelerations, and segment durations, respectively.

\section{MIGHTY} \label{sec:mighty}

\subsection{Decision Variables}
As shown in Sec.~\ref{sec:hermite_spline}, the knot states \(\mathbf{p}_i,\mathbf{v}_i,\mathbf{a}_i (i=0,\dots,M)\) and segment durations \(T_s (s=0,\dots,M-1)\) fully determine the trajectory. 
Thus we optimize the interior positions, velocities, accelerations, and per-segment durations:
\[ \mathbf{z} = 
  \bigl[\, \mathbf{p}_1^{\top},\mathbf{v}_1^{\top},\mathbf{a}_1^{\top},\ldots,\mathbf{p}_{M-1}^{\top},\mathbf{v}_{M-1}^{\top},\mathbf{a}_{M-1}^{\top}, T_0,\ldots,T_{M-1} \bigr]^{\top}. 
\]
with fixed boundary states $(\mathbf{p}_0,\mathbf{v}_0,\mathbf{a}_0)$ and $(\mathbf{p}_M,\mathbf{v}_M,\mathbf{a}_M)$. 

To keep the optimization unconstrained while enforcing certain requirements (\eg $T_s>0$), we apply diffeomorphism, such as
$
  T_s \;=\; \phi(\sigma_s) $ with $\phi(\sigma)=e^{\sigma}, \ \sigma_s\in\mathbb{R}.
$
We also find that optimizing raw derivative knots $(\mathbf{v}_i,\mathbf{a}_i)$ can be numerically unstable.
Sec.~\ref{sec:reparameterizations} introduces scalings that improve stability without changing the optimum. 
For clarity, the formulas below are written in terms of the original variables $(\mathbf{p}_s,\mathbf{v}_s,\mathbf{a}_s,T_s)$; applying a diffeomorphism only inserts chain-rule factors in the gradients.

\subsection{Representations}
Given the quintic Hermite segment in Eq.~\eqref{eq:hermite_segment}, a smooth cost can be expressed in terms of the raw Hermite variables $(\mathbf{p}_s,\mathbf{v}_s,\mathbf{a}_s,\mathbf{p}_{s+1},\mathbf{v}_{s+1},\mathbf{a}_{s+1},T_s)$.
However, evaluating costs and gradients directly in Hermite form can be computationally expensive, and other representations, such as B\'ezier curve, can be more efficient.

For instance, a B\'ezier curve allows efficient evaluation of the trajectory and its derivatives at arbitrary sample points.
A degree-$n$ B\'ezier segment is written as
\(
\mathbf{x}(u)=\sum_{i=0}^{n} B_i^{n}(u)\,\mathbf{c}_i, u\in[0,1],
\)
where the vectors \(\mathbf{c}_i\) are the B\'ezier control points, and the Bernstein basis polynomials are
\(
B_i^{n}(u) \;=\; \binom{n}{i}\, u^{\,i}\,(1-u)^{\,n-i}, i=0,\dots,n.
\)
Because evaluation is just a weighted sum of control points with weights $B_i^n(u)$, sampling reduces to a few dot products with small, reusable tables.
Then each sample is a small dot product:
positions use the $B_i^{5}$ table, and derivatives use the degree-reduced Bernstein weights. 
Since these tables are tiny and shared across iterations, and each dot product has only $6,5,4,3$ terms (for degree $5$ down to $2$), this is faster and numerically steadier than recomputing higher-order Hermite polynomials and their derivatives at every sample.

Although B\'ezier representation has these computational advantages, note that B\'ezier control points do not enforce cross-segment continuity by themselves.
On the other hand, Hermite variables guarantee $\mathcal{C}^k$ continuity across segments by construction. 
Therefore, it is often advantageous to optimize in the Hermite parameterization while evaluating cost terms in the B\'ezier basis to exploit its computational benefits. 
In our simulations and hardware experiments, we optimize over Hermite and evaluate costs in B\'ezier; however, in general, MIGHTY does not require B\'ezier for cost evaluation.

\subsection{Objective and Closed-Form Gradient} \label{sec:grad}

To make the formulation basis-independent, we express the objective and its gradients w.r.t.\ a generic
control-point (or coefficient) vector~$\mathbf{c}$, independent of the chosen parameterization (Hermite end states,
B\'ezier control points, etc.). 
As a concrete example, the affine map between Hermite and B\'ezier on segment $s$ is
\begin{equation}\label{eq:hermite2bezier} 
    \begin{aligned} 
        \mathbf{c}_{s,0} &= \mathbf{p}_s, \ 
        \mathbf{c}_{s,1} = \mathbf{p}_s + \tfrac{T_s}{5}\,\mathbf{v}_s, \ 
        \mathbf{c}_{s,2} = \mathbf{p}_s + \tfrac{2T_s}{5}\,\mathbf{v}_s + \tfrac{T_s^2}{20}\,\mathbf{a}_s, \\ 
        \mathbf{c}_{s,3} &= \mathbf{p}_{s+1} - \tfrac{2T_s}{5}\,\mathbf{v}_{s+1} + \tfrac{T_s^2}{20}\,\mathbf{a}_{s+1}, \ 
        \mathbf{c}_{s,4} = \mathbf{p}_{s+1} - \tfrac{T_s}{5}\,\mathbf{v}_{s+1}, \\ 
        \mathbf{c}_{s,5} &= \mathbf{p}_{s+1}. 
    \end{aligned}
\end{equation}
Let $\mathbf{c}_s=[\mathbf{c}_{s,0}^{\top}\!\cdots\!\mathbf{c}_{s,5}^{\top}]^{\top}$ and $\mathbf{y}_s=[\mathbf{p}_s^{\top},\mathbf{v}_s^{\top},\mathbf{a}_s^{\top},\mathbf{p}_{s+1}^{\top},\mathbf{v}_{s+1}^{\top},\mathbf{a}_{s+1}^{\top}]^{\top}$, so $\mathbf{c}_s=C(T_s)\,\mathbf{y}_s$.
Let the total objective be
\(
J(\mathbf{z}) \;=\; \sum_{s=0}^{M-1} J_s\bigl(\mathbf{c}_s\bigr).
\) 
To obtain $\partial J/\partial \mathbf{z}$, we first compute the derivatives w.r.t $\mathbf{y}_s$ and $T_s$, then apply the chain rule back to $\mathbf{z}$.
Denote
\(
\mathbf{g}_{\mathbf{c}_s} \equiv \partial J_s/\partial \mathbf{c}_s
\)
and the explicit time derivative
\(
\bigl.\partial J_s/\partial T_s|_{\text{explicit}}.
\)
Since ${\partial \mathbf{c}_s}/{\partial \mathbf{y}_s} = C(T_s)$, the chain rule yields
\begin{align}
\label{eq:pullbackY}
\frac{\partial J_s}{\partial \mathbf{y}_s} \;&=\; C(T_s)^\top\,\mathbf{g}_{\mathbf{c}_s},\\[-2pt]
\label{eq:pullbackT}
\frac{\partial J_s}{\partial T_s} \;&=\; \Bigl.\frac{\partial J_s}{\partial T_s}\Bigr|_{\text{explicit}}
\;+\;
\mathbf{g}_{\mathbf{c}_s}^\top \frac{\partial \mathbf{c}_s}{\partial T_s}.
\end{align}
Note that the explicit time derivative
\(
  \bigl.\partial J_s/\partial T_s\bigr|_{\text{explicit}}
\)
comes from factors like $T_s^{5}$ in smoothness integrals.
Using Eq.~\eqref{eq:hermite2bezier}, the entries of $C(T_s)$ are simple constants,
so Eq.~\eqref{eq:pullbackY} expands to the following closed-form expressions:
\begingroup
\small %
\[
\begin{aligned}
\nabla_{\mathbf{p}_s}J_s &= \mathbf{g}_{\mathbf{c}_{s,0}}+\mathbf{g}_{\mathbf{c}_{s,1}}+\mathbf{g}_{\mathbf{c}_{s,2}},
\nabla_{\mathbf{v}_s}J_s = \tfrac{T_s}{5}\,\mathbf{g}_{\mathbf{c}_{s,1}} + \tfrac{2T_s}{5}\,\mathbf{g}_{\mathbf{c}_{s,2}}, \\
\nabla_{\mathbf{a}_s}J_s &= \tfrac{T_s^2}{20}\,\mathbf{g}_{\mathbf{c}_{s,2}},
\nabla_{\mathbf{p}_{s+1}}J_s = \mathbf{g}_{\mathbf{c}_{s,3}}+\mathbf{g}_{\mathbf{c}_{s,4}}+\mathbf{g}_{\mathbf{c}_{s,5}}, \\
\nabla_{\mathbf{v}_{s+1}}J_s &= -\tfrac{2T_s}{5}\,\mathbf{g}_{\mathbf{c}_{s,3}} - \tfrac{T_s}{5}\,\mathbf{g}_{\mathbf{c}_{s,4}},
\nabla_{\mathbf{a}_{s+1}}J_s = \tfrac{T_s^2}{20}\,\mathbf{g}_{\mathbf{c}_{s,3}}.
\end{aligned}
\]
\endgroup

\begin{changeframe}

The coefficient gradient \( \partial \mathbf{c}_s / \partial T_s \) ($=\nabla_{T_s}\mathbf{c}_s$) used in the coefficient path in Eq.~\eqref{eq:pullbackT} can be computed in closed form as
\begin{equation}\label{eq:dJdT_coeff}
  \begin{aligned}
    \nabla_{T_s}\mathbf{c}_{s,0} &= \mathbf{0}, \
    \nabla_{T_s}\mathbf{c}_{s,1} = \tfrac{1}{5}\,\mathbf{v}_s, \
    \nabla_{T_s}\mathbf{c}_{s,2} = \tfrac{2}{5}\,\mathbf{v}_s + \tfrac{T_s}{10}\,\mathbf{a}_s,\\
    \nabla_{T_s}\mathbf{c}_{s,3} &= - \tfrac{2}{5}\,\mathbf{v}_{s+1} + \tfrac{T_s}{10}\,\mathbf{a}_{s+1}, \
    \nabla_{T_s}\mathbf{c}_{s,4} = - \tfrac{1}{5}\,\mathbf{v}_{s+1}, \\
    \nabla_{T_s}\mathbf{c}_{s,5} &= \mathbf{0}.
  \end{aligned}
\end{equation}

\end{changeframe}

Therefore, to obtain the gradient, we need to compute $\mathbf{g}_{\mathbf{c}_s}$ and 
\(
\bigl.\partial J_s/\partial T_s\bigr|_{\text{explicit}}.
\)
Here we present two types of cost terms: (1) cost terms defined on control points, and (2) cost terms defined via sampled states along the trajectory, and their gradients used in Sec.~\ref{sec:simulation-results}.

\subsubsection{Cost on Control Points}\label{par:jerk_cost_and_grad}

We first consider cost terms that are expressed directly in terms of the coefficients/control points, such as integrated squared jerk.
For a quintic B\'ezier segment $s$ with control points $\mathbf{c}_{s,0},\ldots,\mathbf{c}_{s,5}\in\mathbb{R}^3$, obtained from the Hermite endpoint states via Eq.~\eqref{eq:hermite2bezier}, the third forward differences are:
\(
  \Delta_{s,m} = \mathbf{c}_{s,m+3}-3\,\mathbf{c}_{s,m+2}+3\,\mathbf{c}_{s,m+1}-\mathbf{c}_{s,m}, \ m\in\{0,1,2\}. 
\)
Then, the integrated squared jerk on segment \(s\) is
\[
  J_{\text{smooth}, s} \! = \! \int_0^{T_s}\!\!\|\mathbf{j}(t)\|^2\,dt \!=\! C_s \int_0^1 \! \Big\|\sum_{m=0}^{2}\Delta_{s,m}\,B_m^{2}(\tau_s)\Big\|^2 d\tau_s,
\]
where \(C_s = 3600\,T_s^{-5}\).
Define the (scalar) Bernstein-quadratic Gram matrix $G\in\mathbb{R}^{3\times 3}$ by
\(
  G_{mn}\;\triangleq\;\int_{0}^{1} B_m^{2}(\tau_s)\,B_n^{2}(\tau_s)\,d\tau_s
  \quad(m,n=0,1,2),
\)
so that, with $\Delta_s \triangleq [\Delta_{s,0}^\top,\Delta_{s,1}^\top,\Delta_{s,2}^\top]^\top\in\mathbb{R}^9$,
\begin{equation}\label{eq:Jjerk}
  J_{\text{smooth}}
  \;=\;\sum_{s=0}^{M-1} C_s \;\Delta_s^\top\,Q\,\Delta_s,
  \qquad Q \;\triangleq\; G \otimes I_3.
\end{equation}
(Explicitly, $G$ is symmetric positive definite with constants
$G_{00}=G_{22}=\tfrac{1}{5},\; G_{11}=\tfrac{2}{15},\; G_{01}=G_{12}=\tfrac{1}{10},\; G_{02}=\tfrac{1}{30}$.)

Let $\lambda_{\min},\lambda_{\max}>0$ be the smallest and largest eigenvalues of $G$.
Since $Q = G\otimes I_3$, the eigenvalues of $Q$ are the eigenvalues of $G$, each repeated three times. Hence
\(
  \lambda_{\min}\,\|\Delta_s\|^2 \;\le\; \Delta_s^\top Q \Delta_s \;\le\; \lambda_{\max}\,\|\Delta_s\|^2.
\)
Therefore a computationally cheap surrogate is the diagonally weighted form
\(
  \tilde J_{\text{smooth}}
  \;=\;\sum_{s=0}^{M-1} C_s \sum_{m=0}^2 w_m\,\|\Delta_{s,m}\|^2,
\)
with fixed positive weights $w_m$, which is equivalent to the exact cost up to a constant factor bounded by
$\lambda_{\min}$ and $\lambda_{\max}$.

Now we derive \(\partial J_{\text{smooth}}/\partial \mathbf{c}\):
Each $\Delta_{s,m}$ is an affine combination of four consecutive control points with
coefficients $\alpha_{j,m}\in\{-1,3,-3,1\}$ at indices $j=m,\ldots,m+3$.
Hence,
\(
    \partial J_{\text{smooth}}/{\partial \Delta_{s,m}} = 2\,C_s\,\Delta_{s,m}
\)
\(
    \partial J_{\text{smooth}}/{\partial \mathbf{c}_{s,j}} = 2\,C_s \sum_{m=0}^{2} \alpha_{j,m}\,\Delta_{s,m}.
\)
Stacking, we obtain
\begin{equation}\label{eq:jerk_wrt_c}
  \mathbf{g}_{\mathbf{c}_s}
  = \frac{\partial J_{\text{smooth}}}{\partial \mathbf{c}_{s}}
  = \big[(\frac{\partial J_{\text{smooth}}}{\partial \mathbf{c}_{s,0}})^\top \cdots (\frac{\partial J_{\text{smooth}}}{\partial \mathbf{c}_{s,5}})^\top \big]^\top .
\end{equation}
We can now derive \(\partial J_{\text{smooth}}/\partial T_s\).
Eq.~\eqref{eq:pullbackT} has two contributions. 
Since \(C_s = 3600\,T_s^{-5}\), the first contribution is
\begin{align}\label{eq:dJdT_explicit}
  \Bigl.\frac{\partial J_s}{\partial T_s}\Bigr|_{\text{explicit}} = -\frac{5}{T_s}\,J_{{\mathrm{smooth}},s}.
\end{align}
Using Eqs.~\eqref{eq:dJdT_coeff} and \eqref{eq:jerk_wrt_c}, we derive the second term of Eq.~\eqref{eq:pullbackT} and obtain a closed-form expression.

\subsubsection{Cost on Sampled States}\label{sec:sampled_cost_and_grad}

Now we consider cost terms that are evaluated on sampled states of the trajectory.
Fix a set of sample points $\{\tau_{s,j}\}_{j=0}^{\kappa_s}\subset[0,1]$ for segment $s$.
Define a generic per-sample cost $\ell:\mathbb{R}^3\times\mathbb{R}^3\times\mathbb{R}^3\times\mathbb{R}^3\times\mathbb{R}_{>0}\to\mathbb{R}$,
$\ell(\mathbf{x},\mathbf{v},\mathbf{a},\mathbf{j};T)$.
The sampled cost on segment $s$ is
\[
  J_{\text{samp}, s}
  \;=\;
  \sum_{j=0}^{\kappa_s}
  \ell\!\bigl(\mathbf{x}_s(\tau_{s,j}),
               \mathbf{v}_s(\tau_{s,j}),
               \mathbf{a}_s(\tau_{s,j}),
               \mathbf{j}_s(\tau_{s,j});\,T_s\bigr).
\]
Let the sample Jacobians at $\tau_{s,j}$ be
$\ell_{\mathbf{x}}=\partial\ell/\partial\mathbf{x}$,
$\ell_{\mathbf{v}}=\partial\ell/\partial\mathbf{v}$,
$\ell_{\mathbf{a}}=\partial\ell/\partial\mathbf{a}$,
$\ell_{\mathbf{j}}=\partial\ell/\partial\mathbf{j}$. 
Typically $\ell$ has no explicit $T_s$ (as above), so $T_s$ appears only via time-scaling and quadrature weights. \footnote{If an explicit time term is desired, then $\ell=\ell(\mathbf{x},\mathbf{v},\mathbf{a},\mathbf{j};T_s)$ and an extra $\partial \ell/\partial T_s$ term would be added when differentiating w.r.t.\ $T_s$.}
Then, holding $T_s$ fixed, the gradient w.r.t.\ the $k$th control point is
\[
\begin{aligned}
  \mathbf{g}_{c_s,k}
  &=
  \sum_{j=0}^{\kappa_s}\,
  \Bigl[
    B_k^5\,\ell_{\mathbf{x}} \nonumber
    + \tfrac{5}{T_s}\bigl(B_{k-1}^4-B_k^4\bigr)\,\ell_{\mathbf{v}} \\ \nonumber
    &+ \tfrac{20}{T_s^2}\bigl(B_{k-2}^3-2B_{k-1}^3+B_k^3\bigr)\,\ell_{\mathbf{a}} \\ \nonumber
    &+ \tfrac{60}{T_s^3}\bigl(B_{k-3}^2-3B_{k-2}^2+3B_{k-1}^2 -B_k^2\bigr)\,\ell_{\mathbf{j}} 
  \Bigr], \nonumber
\end{aligned}
\]
where we omit the argument of the Bernstein basis polynomials $B(\tau_{s,j})$ for clarity. 
Note that $B_r^n(\cdot)=0$ if $r \notin \{0,\dots,n\}$.
Stacking over $k=0,\ldots,5$ gives $\mathbf{g}_{\mathbf{c}_s}=\partial J_{\text{samp},s}/\partial\mathbf{c}_s$, which is then pulled back (chain rule) and accumulated into the Hermite variables via Eq.~\eqref{eq:pullbackY}.

Changing $T_s$ affects the sampled cost in two ways. 
First, (a) state-scaling: Even if we hold the B\'ezier control points fixed (\ie keep the curve in normalized time $\tau$ unchanged), rescaling $T_s$ changes the physical derivatives via \(t = t_{s} + \tau_s\,T_s\).
Second, (b) coefficients: In Hermite parameterizations the control points themselves depend on $T_s$ via the Hermite$\to$B\'ezier map.
The two items below compute the partial derivative $\partial J_{\text{samp},s}/\partial T_s$ for both contributions.
\begin{itemize}
\item[(a)] State-scaling (fix $\mathbf{c}_s$).
Holding $\mathbf{c}_s$ fixed means $\mathbf{x}(\tau)$ is unchanged at each sample $\tau_{s,j}$, but
\(
\frac{\partial \mathbf{v}}{\partial T_s}=-\frac{1}{T_s}\mathbf{v},\quad
\frac{\partial \mathbf{a}}{\partial T_s}=-\frac{2}{T_s}\mathbf{a},\quad
\frac{\partial \mathbf{j}}{\partial T_s}=-\frac{3}{T_s}\mathbf{j}.
\)
Thus,
\[ \hspace*{-0.25in}
\Bigl.\frac{\partial J_{\text{samp}, s}}{\partial T_s}\Bigr|_{\text{states}} = \sum_{j=0}^{\kappa_s}\, \Bigl[ - \frac{1}{T_s}\,\ell_{\mathbf{v}}\mathbf{v} - \frac{2}{T_s}\,\ell_{\mathbf{a}}\mathbf{a} - \frac{3}{T_s}\,\ell_{\mathbf{j}}\mathbf{j} \Bigr]_{\tau=\tau_{s,j}}. \]

\item[(b)] Coefficient.
In Hermite parameterizations, the B\'ezier control points depend on $T_s$ (\eg tangent/curvature control points include factors of $T_s$ and $T_s^2$). Using Eq.~\eqref{eq:pullbackT} gives
\(
\Bigl.\frac{\partial J_{\text{samp},s}}{\partial T_s}\Bigr|_{\text{coeff}}
=
\mathbf{g}_{c_s}^\top\,\frac{\partial \mathbf{c}_s}{\partial T_s}.
\)
\end{itemize}

Thus, combining (a) \& (b) above yields,
\(
\frac{\partial J_{\text{samp},s}}{\partial T_s}
=
\Bigl.\frac{\partial J_{\text{samp},s}}{\partial T_s}\Bigr|_{\text{states}}
+
\Bigl.\frac{\partial J_{\text{samp},s}}{\partial T_s}\Bigr|_{\text{coeff}}.
\)
Note that since generally sampling cost $\ell$ does not depend on $T_s$ explicitly, \(\partial J_s/{\partial T_s}|_{\text{explicit}}=0\).

\subsection{Reparameterizations for Derivative Variables}\label{sec:reparameterizations}
Each segment with duration $T_s$ is parameterized by normalized time $\tau_s\in[0,1]$; by the chain rule,
\[
\mathbf{v}(t)=\frac{\partial \mathbf{x}}{\partial t}
=\frac{1}{T_s}\frac{\partial \mathbf{x}}{\partial \tau_s},\qquad
\mathbf{a}(t)=\frac{\partial^2 \mathbf{x}}{\partial t^2}
=\frac{1}{T_s^2}\frac{\partial^2 \mathbf{x}}{\partial \tau_s^2}.
\]
Optimizing raw knot derivatives $\{\mathbf{v}_i,\mathbf{a}_i\}$ together with $\{T_s\}$ introduces explicit $1/T_s$ and $1/T_s^2$ factors in the gradients, so short segments have larger influence.
We therefore optimize scaled knot derivatives. 
Define an averaged local time as
\[
  \bar T_i=
  \begin{cases}
  T_0, & i=0,\\ 
  \frac{T_{i-1}+T_i}{2}, & i=1,\dots,M-1,\\
  T_{M-1}, & i=M,
  \end{cases}
\]
and set
\(
  \hat{\mathbf{v}}_i=\bar T_i\,\mathbf{v}_i,\quad
  \hat{\mathbf{a}}_i=\bar T_i^{2}\,\mathbf{a}_i.
\)
\changetext{This reparameterization aims to reduce the explicit $1/T_s$ and $1/T_s^2$ factors' effect in the gradients w.r.t. the derivative decision variables by optimizing the scaled variables $\{\hat{\mathbf{v}}_i,\hat{\mathbf{a}}_i\}$ instead of $\{\mathbf{v}_i,\mathbf{a}_i\}$. At the endpoints, the scaling uses the adjacent segment duration (no averaging), while at interior knots we use the local average duration to balance the influence of the two neighboring segments; in that case, the inverse-duration effects are mitigated but cannot be canceled for both segments simultaneously unless their durations match.}
Our ablation study in Sec.~\ref{subsec:ablation_scaled_vs_unscaled} confirms the expected improvement in performance.

\section{Simulation Results}\label{sec:simulation-results}

All simulations were run on an \texttt{AlienWare Aurora R8} with an Intel\textsuperscript{\textregistered} Core\textsuperscript{TM} i9-9900K CPU and 64\,GB RAM, using Ubuntu 22.04 LTS and ROS~2 Humble.

\subsection{Representation Benchmarking: Simple Case}

\begin{figure}[t]
  \centering
  \includegraphics[width=\columnwidth]{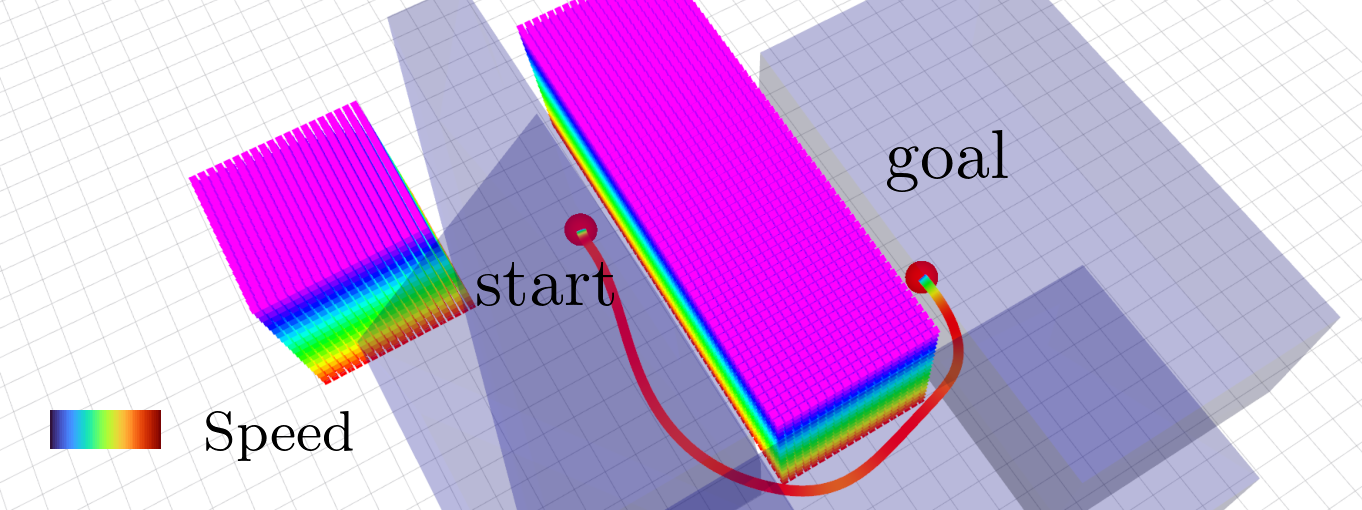}
  \caption{Benchmarking environment for a simple corner avoidance scenario.
  The blue-shaded polygons show the SFC.
  MIGHTY's trajectory is colored by speed (warmer colors indicate faster speeds).}
  \label{fig:simple_benchmarking_rviz}
  \vspace{-0.5em}
\end{figure}

We first benchmark MIGHTY against GCOPTER (which uses MINCO~\cite{wang2022geometrically}) on a simple corner-avoidance task with fixed start and goal (Fig.~\ref{fig:simple_benchmarking_rviz}). 
For a fair comparison, we implement GCOPTER's optimization cost in MIGHTY and use the same optimizer (L-BFGS~\cite{liu1989lbfgs}) and stopping tolerances. 
Both methods start from the same initial guess.

GCOPTER represents each segment as a fifth-order polynomial, so MINCO minimizes jerk (third-order effort) by construction. 
MIGHTY uses a fifth-order Hermite spline to match the polynomial degree. 
GCOPTER applies diffeomorphic parameterizations for waypoints and durations (waypoints remain inside the overlap of safe flight corridors (SFCs); durations stay positive).
The only difference is that MIGHTY includes an explicit jerk smoothness term (Sec.~\ref{par:jerk_cost_and_grad}), while GCOPTER does not add a separate smoothness cost because MINCO already minimizes jerk.

\begin{figure}[t]
  \centering
  \includegraphics[width=\columnwidth]{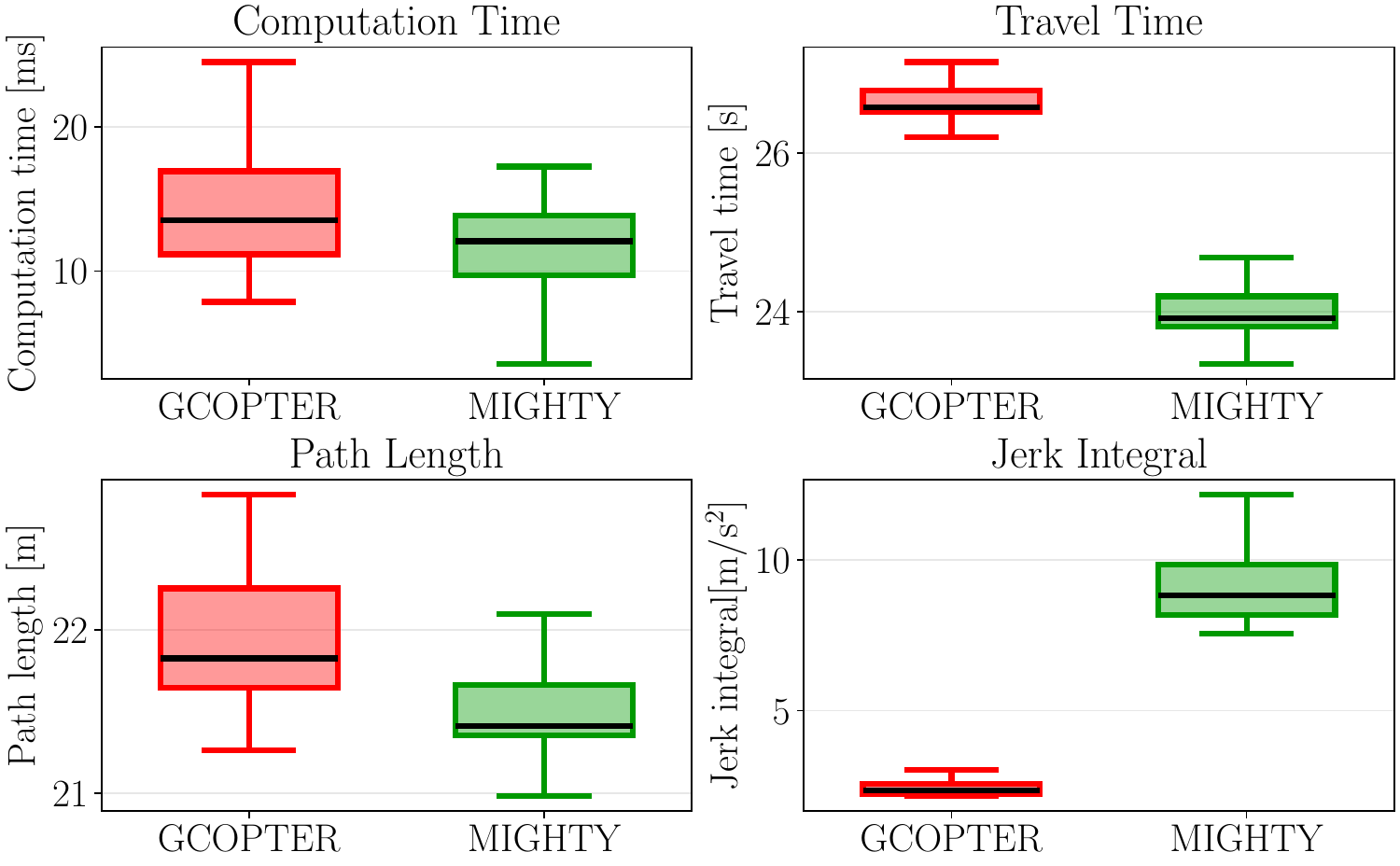}
  \caption{Benchmarking results for MIGHTY and GCOPTER for simple corner avoidance scenario in Fig.~\ref{fig:simple_benchmarking_rviz}. MIGHTY achieves lower computation time, travel time, and path length, while GCOPTER yields lower jerk.}
  \label{fig:simple_benchmarking}
  \vspace{-0.5em}
\end{figure}

We enforce max velocity, $v_{\max}=\SI{1.0}{\m/\s}$, max tilt angle, $\theta_{\max}=\SI{1.05}{rad}$ ($\approx\SI{60}{\degree}$), max angular velocity, $\omega_{\max}=\SI{2.1}{rad/\s}$ ($\approx\SI{120}{\degree/\s}$), max $f_{\max}=\SI{12.0}{\newton}$ and min $f_{\min}=\SI{2.0}{\newton}$ thrust.
We set $w_T=\num{e2}$, $w_{\text{SFC}}=\num{e4}$, $w_v=\num{e4}$, $w_{\theta}=\num{e4}$, $w_{\omega}=\num{e4}$, and $w_f=\num{e5}$ for both MIGHTY and GCOPTER, where $w_T$, $w_{\text{SFC}}$, $w_v$, $w_{\theta}$, $w_{\omega}$, and $w_f$ are weights for time, SFC, velocity, angle, angular velocity, and thrust penalties, respectively. 
These match GCOPTER's default values in their code except $w_T$ (default $20$). 
We use $w_T=\num{e2}$ because $w_T=\num{e3}$ caused occasional constraint violations ($>1\%$) in GCOPTER.
For MIGHTY, we set the jerk smoothness weight to $w_{\mathrm{smooth}}=\num{e-5}$.
GCOPTER sets yaw to the direction of motion, so yaw and yaw rate are determined by $\mathbf{p}$. 
Accordingly, although neither GCOPTER nor MIGHTY optimizes yaw directly, both penalize tilt angle and angular velocity. 
We configure MIGHTY this way solely for fair comparison with GCOPTER. 
Outside this benchmark (see Sec.~\ref{subsec:benchmarking_static_envs}), MIGHTY can use a different cost design.

The result in Fig.~\ref{fig:simple_benchmarking} shows that
MIGHTY achieves a shorter travel time, with slightly faster computation and shorter path than GCOPTER. 
While GCOPTER produces smoother trajectories with lower jerk due to its constraint within the MINCO class, MIGHTY explores a larger search space to identify faster and more effective trajectories. 
This flexibility enables MIGHTY to achieve superior performance, \changetext{and Sec.~\ref{par:smooth_weight_sweep_benchmarking} further investigates the jerk-performance trade-off.}

\subsection{Smooth Weight Sweep Benchmarking}\label{par:smooth_weight_sweep_benchmarking}

\begin{figure}[t]
  \begin{changeframe}
  \centering
  \includegraphics[width=\columnwidth]{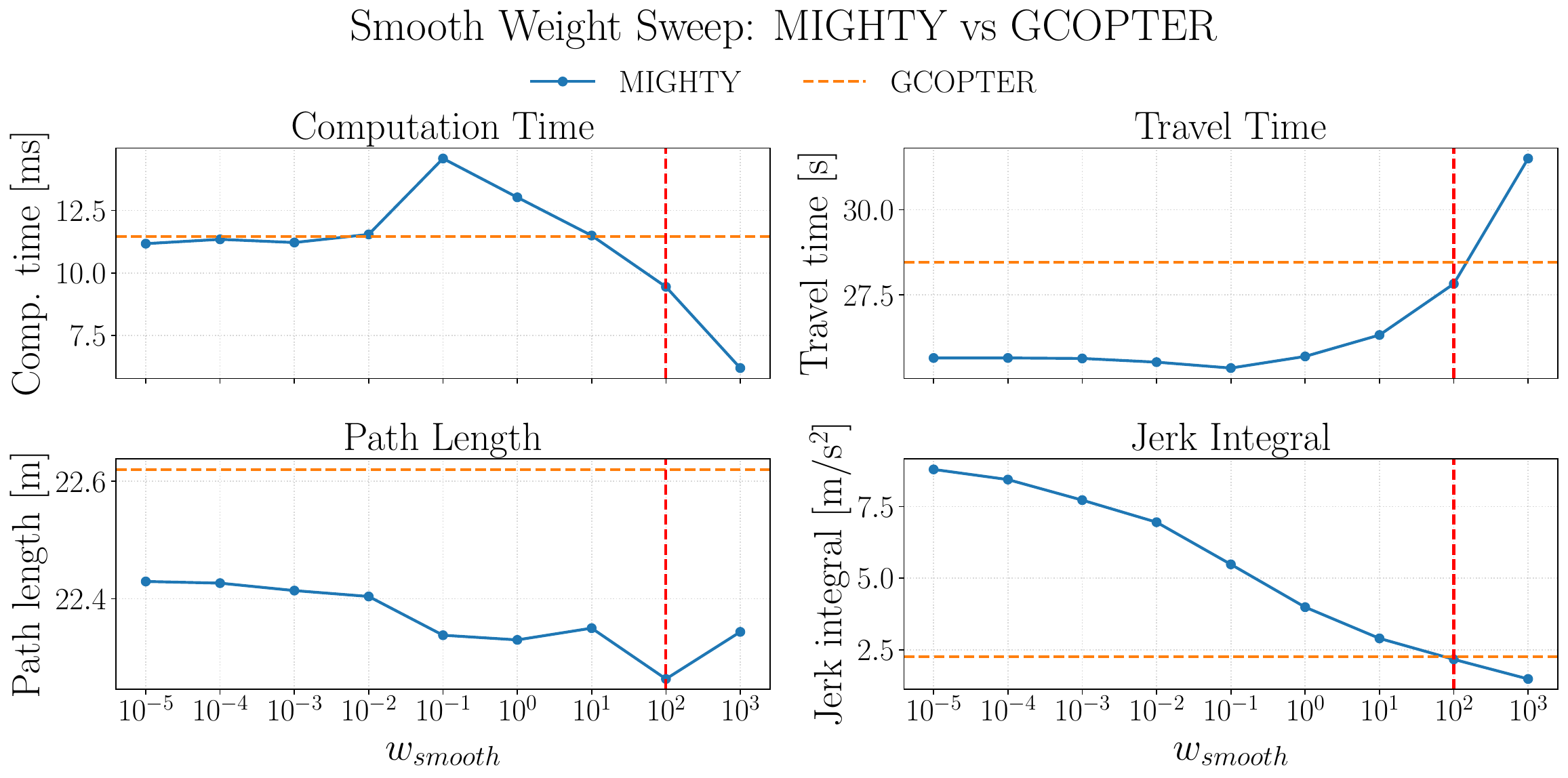}
  \end{changeframe}
  \caption{\changetext{Effect of jerk-smoothness weighting in the simple corner-avoidance scenario (Fig.~\ref{fig:simple_benchmarking_rviz}).
  As $w_{\mathrm{smooth}}$ increases, MIGHTY's jerk decreases toward GCOPTER's jerk (orange dotted reference).
  At $w_{\mathrm{smooth}}=\num{e2}$ (vertical red line), MIGHTY matches GCOPTER's jerk with comparable performance and lower computation time.}}
  \label{fig:smooth_weight_sweep_benchmarking}
  \vspace{-0.5em}
\end{figure}

\changetext{%
Fig.~\ref{fig:simple_benchmarking} shows MIGHTY yields higher jerk than GCOPTER while achieving better travel time.
To quantify this trade-off, we sweep $w_{\mathrm{smooth}}\in\{\num{e-5},\ldots,\num{e3}\}$ in the same scenario.
Fig.~\ref{fig:smooth_weight_sweep_benchmarking} shows that as $w_{\mathrm{smooth}}$ increases, MIGHTY progressively reduces jerk, eventually surpassing GCOPTER's levels of smoothness.
At $w_{\mathrm{smooth}}=\num{e2}$, MIGHTY matches GCOPTER's jerk while maintaining better performance and lower computation time, demonstrating that MIGHTY's higher jerk is a tunable design choice rather than an inherent limitation.
}

\subsection{Local Control of Higher-Order Dynamics}\label{subsec:ref_benchmarking}

\changetext{%
MINCO-based planners lack direct parameterization of higher-order derivatives at individual knots; these quantities are derived from a globally coupled coefficient system (Sec.~\ref{subsec:related_work}).
While GCOPTER can add derivative constraints as soft penalties, MIGHTY's Hermite parameterization treats knot positions and velocities as explicit optimization variables.
To evaluate this difference, we extend the simple corner-avoidance scenario (Fig.~\ref{fig:simple_benchmarking_rviz}) with soft reference tracking terms for position and velocity at the penultimate knot, which is relevant for waypoint velocity specification and hover-at-goal behaviors.
}

\begin{centering}
\begin{figure}[t]
  \begin{changeframe}
  \centering
  \includegraphics[width=\columnwidth]{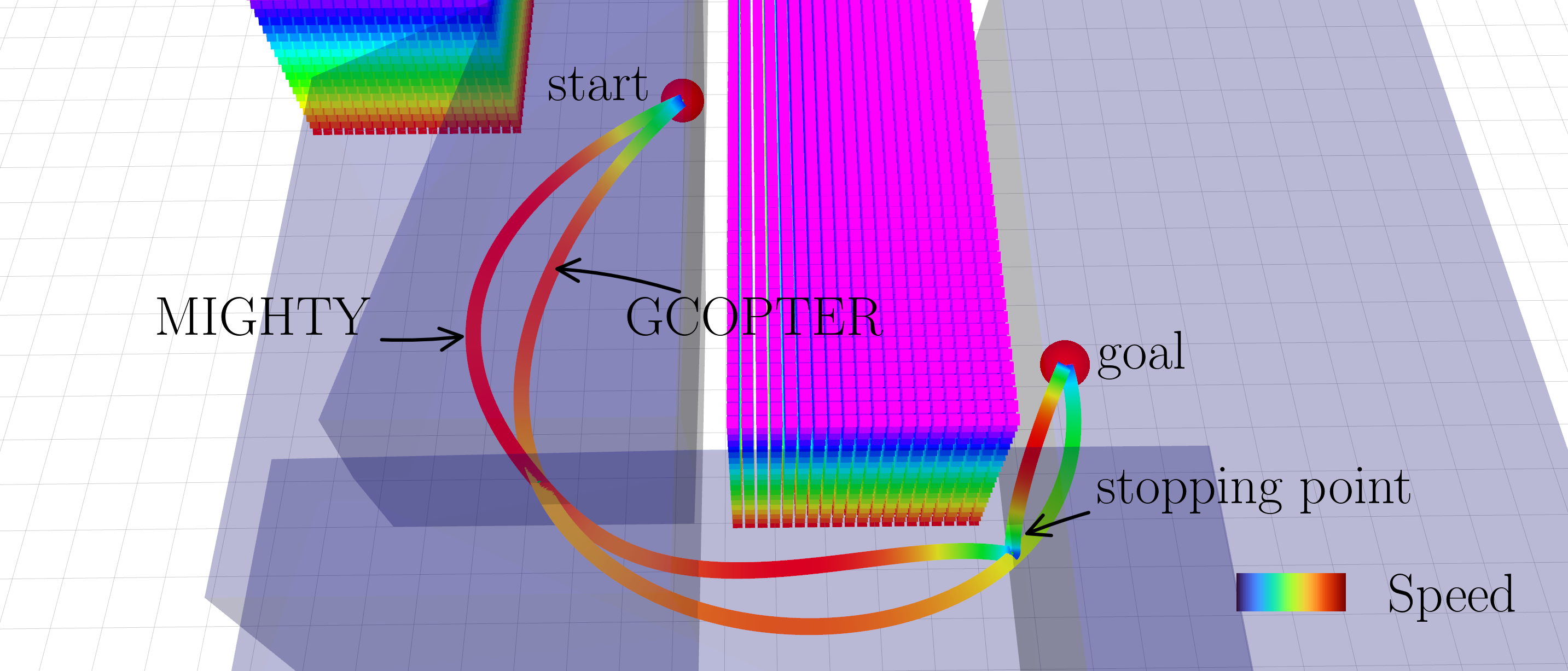}
  \end{changeframe}
  \caption{\changetext{Local control of higher-order dynamics: position and velocity reference tracking.
  GCOPTER and MIGHTY trajectories are color-mapped by speed.
  Both planners enforce a velocity reference $\mathbf{v}_{\text{ref}}=(0,0,0)\,\si{m/s}$ and position reference at the penultimate knot (one knot before the goal).
  As shown, MIGHTY achieves better reference tracking accuracy.
  }}
  \label{fig:ref_benchmarking_rviz}
  \vspace{-0.5em}
\end{figure}
\end{centering}

\changetext{%
We augment the objective function with soft position and velocity reference terms
\[
J_{\text{ref}} = \tfrac{1}{2} w_{\mathbf{p}_{\text{ref}}} \lVert \mathbf{p}(t_k) - \mathbf{p}_{\text{ref}} \rVert^2
              + \tfrac{1}{2} w_{\mathbf{v}_{\text{ref}}} \lVert \mathbf{v}(t_k) - \mathbf{v}_{\text{ref}} \rVert^2,
\]
where $\mathbf{p}_{\text{ref}}$ and $\mathbf{v}_{\text{ref}}$ are desired position and velocity at knot $k$, and $w_{\mathbf{p}_{\text{ref}}}$ and $w_{\mathbf{v}_{\text{ref}}}$ are penalty weights.
For this benchmark, we set $k$ to the penultimate knot (one knot before the goal), $\mathbf{v}_{\text{ref}}=(0,0,0)\,\si{m/s}$ to enforce near-hover at the goal, and $\mathbf{p}_{\text{ref}}$ to the goal position.
We use $w_{\mathbf{p}_{\text{ref}}}=\num{1e3}$ and $w_{\mathbf{v}_{\text{ref}}}=\num{1e3}$, with $v_{\max}=\SI{2.0}{m/s}$ and $w_{\mathrm{smooth}}=\num{e-5}$ for MIGHTY.
All other settings match those in Sec.~\ref{subsec:benchmarking_static_envs}.
}
\changetext{%
We evaluate both GCOPTER and MIGHTY across 100 runs in the same simple corner-avoidance environment.
Fig.~\ref{fig:ref_benchmarking_rviz} shows example trajectories from both planners, and Table~\ref{tab:ref_benchmarking} summarizes the results.
}

\begin{table}[t]
  \centering
  \caption{\changetext{Local control of higher-order dynamics benchmark. We added position/velocity soft penalties on both GCOPTER and MIGHTY and evaluated performance. Best results are in \best{green}.}}
  \label{tab:ref_benchmarking}
  \begin{changeframe}
  \renewcommand{\arraystretch}{1.2}
  \centering
  \resizebox{0.95\columnwidth}{!}{
  \begin{tabular}{l cccc}
    \toprule
    \textbf{Planner} & $e_{\mathbf{p}_{\text{ref}}}$ [m] & $e_{\mathbf{v}_{\text{ref}}}$ [m/s] & $T_{\mathrm{trav}}$ [s] & $T_{\mathrm{comp}}$ [ms] \\
    \midrule
    GCOPTER & 0.017 & 1.0 & 19.7 & \best{7.4} \\
    MIGHTY  & \best{0.0041} & \best{0.16} & \best{14.8} & 10.6 \\
    \bottomrule
  \end{tabular}
  }
  \end{changeframe}
  \vspace{-0.5em}
\end{table}

\changetext{%
MIGHTY achieves substantially better tracking: lower position error (\SI{0.0041}{m} vs. \SI{0.017}{m}), \best{6.5$\times$} lower velocity error (\SI{0.16}{m/s} vs. \SI{1.0}{m/s}), and \best{25.1\%} shorter travel time (\SI{14.8}{s} vs. \SI{19.7}{s}).
GCOPTER is faster (\best{\SI{7.4}{ms}} vs. \SI{10.6}{ms}), though both meet real-time budgets.
This demonstrates MIGHTY's advantage: while GCOPTER adds derivative constraints as soft penalties on MINCO-derived velocities, MIGHTY directly optimizes knot positions and velocities as explicit Hermite variables, enabling superior local control of higher-order dynamics.
}

\subsection{Representation Benchmarking: Complex Case}

\begin{figure}[t]
  \centering
  \includegraphics[width=\columnwidth]{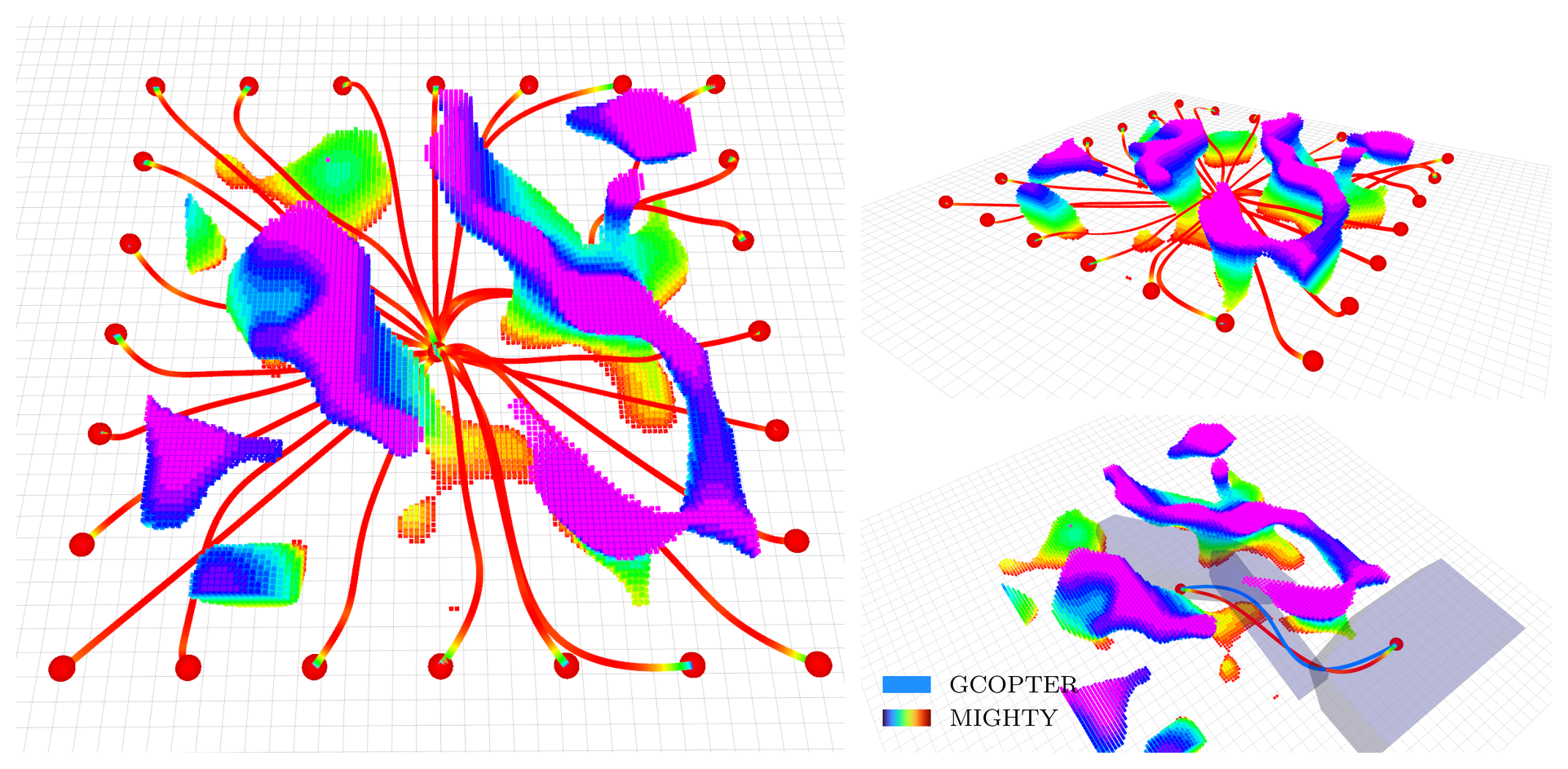}
  \caption{Complex-scene benchmarking setup. MIGHTY's trajectory is color-mapped by speed (warm=fast), and GCOPTER's trajectory is blue.
  The blue-shaded polygons show the SFC.
  The start is at $(0,0,0.5)\,\si{m}$, and 24 goals lie on a grid with $x,y\in[-15,15]\,\si{m}$ and $z=\SI{2.5}{m}$.}
  \label{fig:complex_benchmarking_combined}
  \vspace{-0.5em}
\end{figure}

\begin{table}[t]
\setlength{\heavyrulewidth}{1.0pt}  %
\setlength{\lightrulewidth}{1.0pt}  %
\centering
  \caption{Complex-scene benchmark across 24 goals and five speed limits ($v_{\max}\!\in\!\{1,2, 3, 4 ,5\}$\,m/s). GC = GCOPTER, MI = MIGHTY.}
  \label{tab:bench_complex}
  \renewcommand{\arraystretch}{1.0}
  \resizebox{\columnwidth}{!}{
  \begin{tabular}{
    >{\centering\arraybackslash}m{0.1\columnwidth}
    >{\centering\arraybackslash}m{0.1\columnwidth}
    >{\centering\arraybackslash}m{0.2\columnwidth}
    >{\centering\arraybackslash}m{0.15\columnwidth}
    >{\centering\arraybackslash}m{0.18\columnwidth}
    >{\centering\arraybackslash}m{0.15\columnwidth}
    >{\centering\arraybackslash}m{0.15\columnwidth}
  }
    \toprule
    $v_{\max}$ [m/s] & \textbf{Algo.} & $T_{\mathrm{comp}}$ [ms] & $T_{\mathrm{trav}}$ [s] & $L_{\mathrm{path}}$ [m] & $\int \lVert \mathbf{j}(t) \rVert \, dt$ [m/s$^{2}$] & $\rho_{\mathrm{viol}}$ [\%] \\
    \midrule
    \multirow{3}{*}{1.0}
      & GC & \best{7.7} & 25.1 & 19.1 & \best{2.1} & \best{0.0} \\
      \cline{2-7}
      & MI 
        & \begin{tabular}[c]{@{}c@{}}8.1\\[-1ex] {\scriptsize (+4.8\%)}\end{tabular}
        & \begin{tabular}[c]{@{}c@{}}\best{21.5}\\[-1ex] {\scriptsize (\best{-14.2\%})}\end{tabular}
        & \begin{tabular}[c]{@{}c@{}}\best{18.8}\\[-1ex] {\scriptsize (\best{-1.4\%})}\end{tabular}
        & \begin{tabular}[c]{@{}c@{}}6.9\\[-1ex] {\scriptsize (+229.0\%)}\end{tabular}
        & \begin{tabular}[c]{@{}c@{}}\best{0.0}\\[-1ex] {\scriptsize (0.0\%)}\end{tabular} \\
    \midrule
    \multirow{3}{*}{2.0}
      & GC & 9.8 & 12.6 & 19.0 & \best{7.9} & \best{0.0} \\
      \cline{2-7}
      & MI 
        & \begin{tabular}[c]{@{}c@{}}\best{9.3}\\[-1ex] {\scriptsize (\best{-4.7\%})}\end{tabular}
        & \begin{tabular}[c]{@{}c@{}}\best{11.0}\\[-1ex] {\scriptsize (\best{-12.8\%})}\end{tabular}
        & \begin{tabular}[c]{@{}c@{}}\best{18.6}\\[-1ex] {\scriptsize (\best{-2.2\%})}\end{tabular}
        & \begin{tabular}[c]{@{}c@{}}17.2\\[-1ex] {\scriptsize (+118.5\%)}\end{tabular}
        & \begin{tabular}[c]{@{}c@{}}\best{0.0}\\[-1ex] {\scriptsize (0.0\%)}\end{tabular} \\
    \midrule
    \multirow{3}{*}{3.0}
      & GC & 11.8 & 8.5 & 19.0 & \best{13.9} & \best{0.0} \\
      \cline{2-7}
      & MI 
        & \begin{tabular}[c]{@{}c@{}}\best{7.8}\\[-1ex] {\scriptsize (\best{-33.6\%})}\end{tabular}
        & \begin{tabular}[c]{@{}c@{}}\best{7.7}\\[-1ex] {\scriptsize (\best{-9.8\%})}\end{tabular}
        & \begin{tabular}[c]{@{}c@{}}\best{18.7}\\[-1ex] {\scriptsize (\best{-1.4\%})}\end{tabular}
        & \begin{tabular}[c]{@{}c@{}}27.2\\[-1ex] {\scriptsize (+95.9\%)}\end{tabular}
        & \begin{tabular}[c]{@{}c@{}}\best{0.0}\\[-1ex] {\scriptsize (0.0\%)}\end{tabular} \\
    \midrule
    \multirow{3}{*}{4.0}
      & GC & 7.7 & 6.9 & 18.8 & \best{17.6} & \best{0.0} \\
      \cline{2-7}
      & MI 
        & \begin{tabular}[c]{@{}c@{}}\best{7.2}\\[-1ex] {\scriptsize (\best{-5.5\%})}\end{tabular}
        & \begin{tabular}[c]{@{}c@{}}\best{6.0}\\[-1ex] {\scriptsize (\best{-13.0\%})}\end{tabular}
        & \begin{tabular}[c]{@{}c@{}}\best{18.6}\\[-1ex] {\scriptsize (\best{-1.2\%})}\end{tabular}
        & \begin{tabular}[c]{@{}c@{}}35.6\\[-1ex] {\scriptsize (+102.6\%)}\end{tabular}
        & \begin{tabular}[c]{@{}c@{}}\best{0.0}\\[-1ex] {\scriptsize (0.0\%)}\end{tabular} \\
    \midrule
    \multirow{3}{*}{5.0}
      & GC & \best{6.8} & 6.1 & 18.8 & \best{19.8} & \best{0.0} \\
      \cline{2-7}
      & MI 
        & \begin{tabular}[c]{@{}c@{}}7.2\\[-1ex] {\scriptsize (+5.7\%)}\end{tabular}
        & \begin{tabular}[c]{@{}c@{}}\best{5.3}\\[-1ex] {\scriptsize (\best{-14.1\%})}\end{tabular}
        & \begin{tabular}[c]{@{}c@{}}\best{18.7}\\[-1ex] {\scriptsize (\best{-0.7\%})}\end{tabular}
        & \begin{tabular}[c]{@{}c@{}}43.7\\[-1ex] {\scriptsize (+120.5\%)}\end{tabular}
        & \begin{tabular}[c]{@{}c@{}}\best{0.0}\\[-1ex] {\scriptsize (0.0\%)}\end{tabular} \\
    \midrule\midrule
    \multicolumn{2}{c}{\textbf{Overall}} & \best{-9.3\%} & \best{-13.1\%} & \best{-1.4\%} & {+113.3\%} & {0.0\%} \\
    \bottomrule
  \end{tabular}}
  \vspace{-0.5em}
\end{table}

We further evaluate MIGHTY in a complex environment (Fig.~\ref{fig:complex_benchmarking_combined}) with start at $(0,0,0.5)\,\si{m}$ and 24 goals at $x,y\in[-15,15]\,\si{m}$ and $z=2.5\,\si{m}$. 
Optimization settings are identical to the simple case for both methods.
We report the following metrics:
$v_{\max}$ [m/s]: Speed limit,
$T_{\mathrm{comp}}$ [ms]: Computation time,
$T_{\mathrm{trav}}$ [s]: Travel time,
$L_{\mathrm{path}}$ [m]: Path length,
$\int \lVert \mathbf{j}(t) \rVert \, dt$ [m/s$^{2}$]: Jerk smoothness integral, and 
$\rho_{\mathrm{viol}}$ [\%]: Fraction of time violating dynamic bounds.

Overall, Table~\ref{tab:bench_complex} shows MIGHTY reduces computation time by \best{9.3\%}, travel time by \best{13.1\%}, and path length by \best{1.4\%}.
GCOPTER yields lower jerk, while MIGHTY's higher jerk reflects more aggressive maneuvers, \changetext{enabled by smaller $w_{\mathrm{smooth}}=\num{e-5}$ and its larger search space.}
Both planners keep $\rho_{\mathrm{viol}}\le\SI{1.0}{\percent}$.

\subsection{Ablation Study on Scaled and Unscaled Variables}\label{subsec:ablation_scaled_vs_unscaled}
\begin{figure}[t]
  \centering
  \includegraphics[width=\columnwidth]{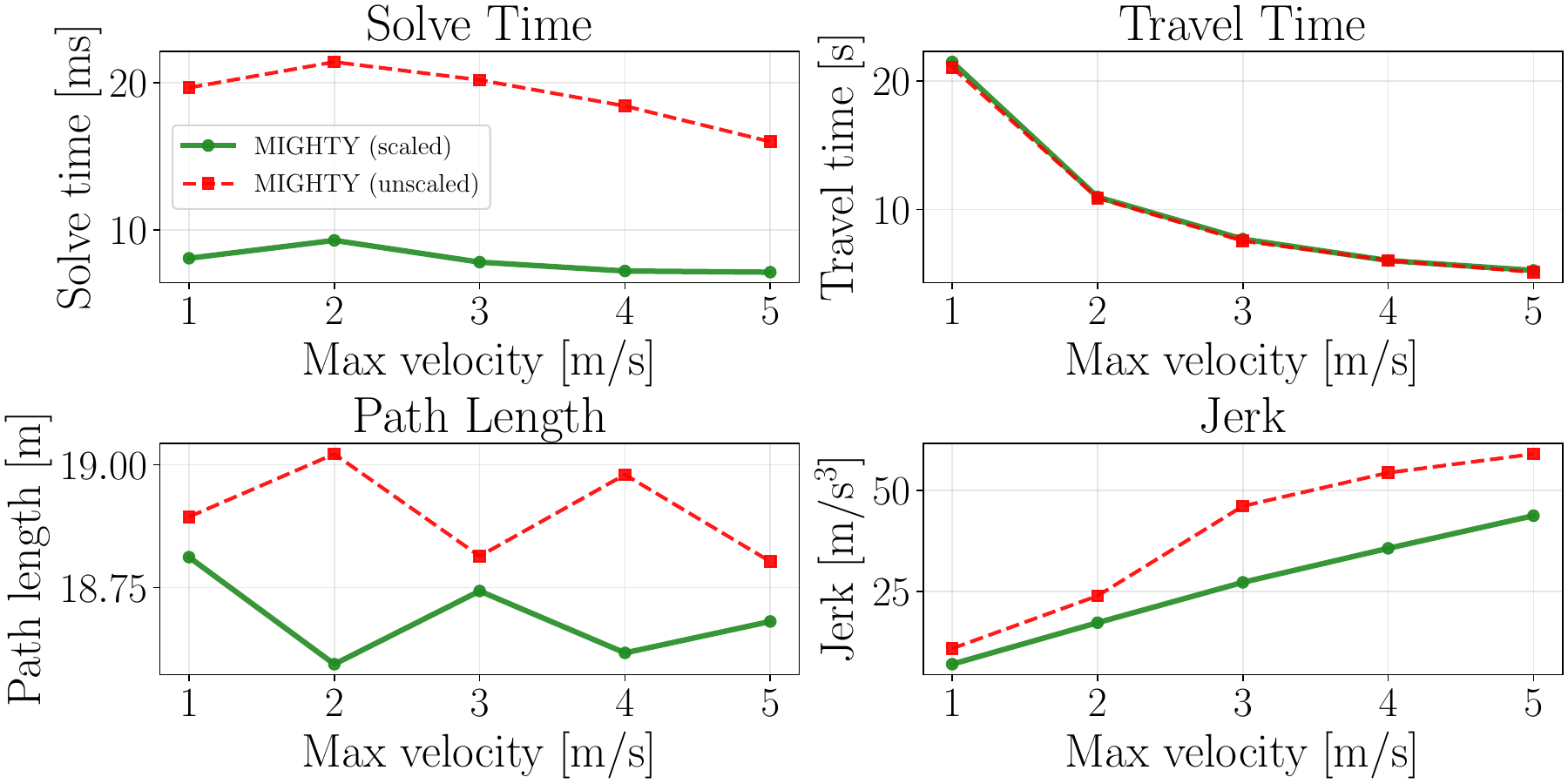}
  \caption{Ablation study comparing MIGHTY with scaled and unscaled variables in the complex benchmarking scenario. The scaled version shows $\approx 2\times$ faster computation time, with lower jerk and slightly shorter path length, while travel time is nearly identical.}
  \label{fig:scaled_vs_unscaled}
  \vspace{-0.5em}
\end{figure}
As discussed in Sec.~\ref{sec:reparameterizations}, we optimize over scaled variables to improve numerical stability.
We evaluate this via ablation using the same complex benchmarking setup.
Fig.~\ref{fig:scaled_vs_unscaled} shows the scaled variant is $\approx 2\times$ faster with lower jerk, while path length and travel time are nearly identical, confirming that scaling improves efficiency without degrading quality.

\subsection{Benchmarking in Static Environments}\label{subsec:benchmarking_static_envs}
\begin{figure}[t]
  \centering
  \includegraphics[width=\columnwidth]{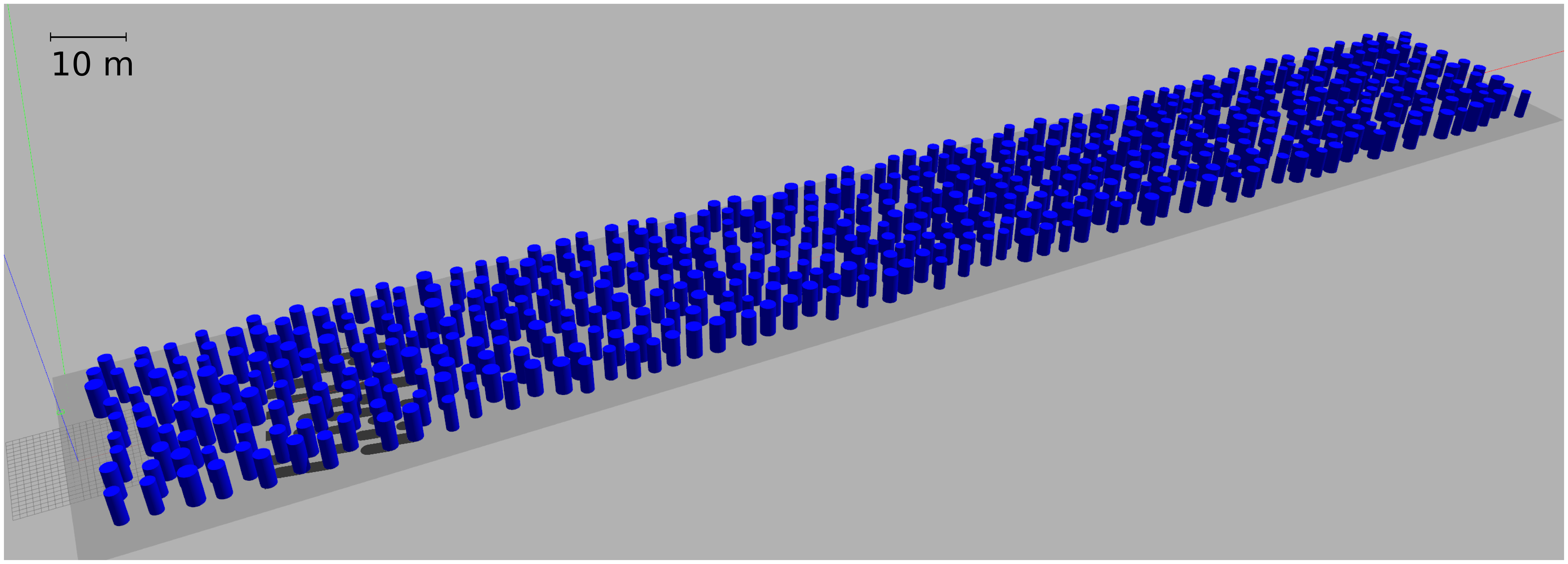}
  \vspace{0.3em}
  \includegraphics[width=\columnwidth]{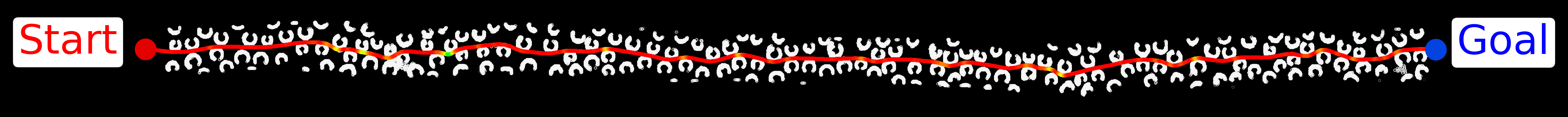}
  \caption{(Top) Static obstacle environment used for benchmarking. 
  (Bottom) Top view of the point cloud and path generated by MIGHTY in the static environment. 
  The warmer colors indicate faster speeds.}
  \label{fig:static_env_and_path}
   \vspace{-0.5em}
\end{figure}%
\begin{table*}
  \caption{Benchmarking results (safety, computation time, performance, and constraint violation).
  We mark in \best{green} the best value in each column \emph{considering only configurations with $R_{\mathrm{succ}}=100\%$}; 
  configurations with $R_{\mathrm{succ}}<100\%$ are excluded from best highlighting.
  For EGO-Swarm2, we evaluated both its default sensor, depth camera, and a LiDAR (as used by SUPER and MIGHTY) and report results as ``camera'' \,|\, ``LiDAR''. 
  We also tested EGO\text{-}Swarm with LiDAR, but its performance was substantially worse, so we omit it from the table for clarity.
  For SUPER, computation time is reported as ``exploratory'' \,\&\, ``safe''.
  The data are shaded red if they have unsuccessful rates $R_{\mathrm{succ}}<100\%$.}
  \label{tab:local_trajectory_optimization_benchmarking}
  \centering
  \renewcommand{\arraystretch}{1.2}
  \resizebox{\textwidth}{!}{
    \begin{tabular}{ 
      c c c c c c c c c c c c c c c
    }
      \toprule
      \multirow{2}{*}[-0.4ex]{\textbf{Algorithm}}
      & \multirow{2}{*}[-0.4ex]{\textbf{Params}}
      & \multirow{2}{*}[-0.4ex]{\textbf{Traj.}}
      & \multicolumn{1}{c}{\textbf{Success}}
      & \multicolumn{2}{c}{\textbf{Computation Time}}
      & \multicolumn{4}{c}{\textbf{Performance}}
      & \multicolumn{3}{c}{\textbf{Constraint Violation}}
      \\
      \cmidrule(lr){4-4} 
      \cmidrule(lr){5-6}
      \cmidrule(lr){7-10}
      \cmidrule(lr){11-13}
      &&
      & $R_{\mathrm{succ}}$ [\%]
      & $T_{\mathrm{opt}}$ [ms]
      & $T_{\mathrm{total}}$ [ms]
      & $T_{\mathrm{trav}}$ [s]
      & $L_{\mathrm{path}}$ [m]
      & $S_{\mathrm{jerk}}$ [m/s$^{2}$]
      & $\bar{S}_{\mathrm{jerk}}$ [m/s$^{3}$]
      & $\rho_{\mathrm{vel}}$ [\%]
      & $\rho_{\mathrm{acc}}$ [\%]
      & $\rho_{\mathrm{jerk}}$ [\%]
      \\
      \midrule
      \multirow{4}{*}{\textbf{EGO-Swarm}} & $v_{\max}=\SI{4.0}{m/s}$ & \multirow{4}{*}{B-spline} & \cellcolor{red!10} \worst{40.0} & \cellcolor{red!10} {1.8} & \cellcolor{red!10} {2.3} & \cellcolor{red!10} {135.9} & \cellcolor{red!10} {379.7} & \cellcolor{red!10} {4233.1} & \cellcolor{red!10} {43.6} & \cellcolor{red!10} {3.5} & \cellcolor{red!10} {6.5} & \cellcolor{red!10} {38.0} \\
      & $v_{\max}=\SI{3.0}{m/s}$ & & \cellcolor{red!10} \worst{70.0} & \cellcolor{red!10} {1.8} & \cellcolor{red!10} {2.4} & \cellcolor{red!10} {178.2} & \cellcolor{red!10} {380.2} & \cellcolor{red!10} {3024.5} & \cellcolor{red!10} {26.7} & \cellcolor{red!10} {5.7} & \cellcolor{red!10} {1.2} & \cellcolor{red!10} {16.5} \\
      & $v_{\max}=\SI{2.0}{m/s}$ & & \cellcolor{red!10} \worst{80.0} & \cellcolor{red!10} {0.9} & \cellcolor{red!10} {1.3} & \cellcolor{red!10} {262.7} & \cellcolor{red!10} {364.8} & \cellcolor{red!10} {1074.9} & \cellcolor{red!10} {7.3} & \cellcolor{red!10} {3.8} & \cellcolor{red!10} {0.01} & \cellcolor{red!10} {0.4} \\
      & $v_{\max}=\SI{1.0}{m/s}$ & & \cellcolor{red!10} \worst{90.0} & \cellcolor{red!10} {0.6} & \cellcolor{red!10} {0.9} & \cellcolor{red!10} {559.8} & \cellcolor{red!10} {370.0} & \cellcolor{red!10} {204.4} & \cellcolor{red!10} {0.6} & \cellcolor{red!10} {0.1} & \cellcolor{red!10} {0.0} & \cellcolor{red!10} {0.0} \\
      \midrule
      \multirow{2}{*}{\textbf{EGO-Swarm2}} & $w_{\rm obst}=\num{1e+4}$ & \multirow{2}{*}{MINCO} & \best{100.0} | \rshade{\worst{20.0}} & {1.3} | \rshade{{5.9}} & {1.8} | \rshade{{7.6}} & {144.7} | \rshade{{144.6}} & {329.8} | \rshade{{331.7}} & {263.3} | \rshade{{336.8}} & {2.5} | \rshade{{3.5}} & \best{0.0} | \rshade{{0.6}} & \best{0.0} | \rshade{{0.0}} & \best{0.0} | \rshade{{0.01}} \\
      & $w_{\rm obst}=\num{1e+5}$ & & \best{100.0} | \rshade{\worst{70.0}} & \best{1.2} | \rshade{{4.6}} & \best{1.7} | \rshade{{5.9}} & {140.2} | \rshade{{150.1}} & {326.9} | \rshade{{333.5}} & {243.2} | \rshade{{247.7}} & {2.2} | \rshade{{3.5}} & {0.05} | \rshade{{0.02}} & \best{0.0} | \rshade{{0.0}} & \best{0.0} | \rshade{{0.0}} \\
      \midrule
      \multirow{4}{*}{\textbf{SUPER}} & $w_t=\num{1e+4}$ & \multirow{4}{*}{MINCO} & \cellcolor{red!10} \worst{20.0} & \cellcolor{red!10} {5.9 \& 18.0} & \cellcolor{red!10} {46.4} & \cellcolor{red!10} {87.5} & \cellcolor{red!10} {323.5} & \cellcolor{red!10} {627.4} & \cellcolor{red!10} {10.9} & \cellcolor{red!10} {0.0} & \cellcolor{red!10} {0.02} & \cellcolor{red!10} {1.0} \\
      & $w_t=\num{1e+3}$ & & \cellcolor{red!10} \worst{60.0} & \cellcolor{red!10} {5.6 \& 18.0} & \cellcolor{red!10} {57.5} & \cellcolor{red!10} {99.6} & \cellcolor{red!10} {323.4} & \cellcolor{red!10} {431.9} & \cellcolor{red!10} {6.7} & \cellcolor{red!10} {0.0} & \cellcolor{red!10} {0.02} & \cellcolor{red!10} {0.2} \\
      & $w_t=\num{1e+2}$ & & \cellcolor{red!10} \worst{60.0} & \cellcolor{red!10} {6.8 \& 18.2} & \cellcolor{red!10} {47.6} & \cellcolor{red!10} {124.7} & \cellcolor{red!10} {326.9} & \cellcolor{red!10} {399.5} & \cellcolor{red!10} {4.2} & \cellcolor{red!10} {0.0} & \cellcolor{red!10} {0.01} & \cellcolor{red!10} {0.06} \\
      & $w_t=\num{1e+1}$ & & \best{100.0} & {7.8 \& 18.9} & {50.6} & {264.7} & {342.0} & \best{200.0} & \best{0.9} & \best{0.0} & \best{0.0} & \best{0.0} \\
      \midrule
      \multicolumn{2}{c}{\textbf{MIGHTY}} & Hermite & \best{100.0} & {10.5} & {19.7} & \best{79.0} & \best{310.9} & {522.2} & {9.4} & \best{0.0} & \best{0.0} & \best{0.0} \\
      \bottomrule
    \end{tabular}
    }
    \vspace{-1.7em}
\end{table*}%

To thoroughly evaluate MIGHTY, we benchmark it in an obstacle-rich environment (Fig.~\ref{fig:static_env_and_path}) against EGO-Swarm~\cite{zhou2021ego-swarm}, EGO-Swarm2~\cite{zhou2022swarm}, and SUPER~\cite{ren2025super}.
Static cylindrical obstacles are placed randomly with radii sampled in $[\,\SI{1.0}{m},\SI{1.5}{m}\,]$ and height \SI{6.0}{m}, occupying a $\SI{300}{m}\times\SI{40}{m}$ area.
The agent starts at $(0,0,3)\,\si{m}$ and the goal is $(305,0,3)\,\si{m}$.
We constrain velocity, acceleration, and jerk to $v_{\max}=\SI{4.0}{m/s}$, $a_{\max}=\SI{10.0}{m/s^2}$, and $j_{\max}=\SI{30.0}{m/s^3}$.
\changetext{For global planning we run A*, then build an SFC~\cite{liu2017planning} for local optimization.
The intermediate waypoints for A* are projected from the terminal goal onto the local occupancy map centered on the drone.
The average computation time for A* and SFC generation is \SI{0.26}{ms} and \SI{0.48}{ms}, respectively.}

For soft constraints we use a smooth hinge $\phi_\mu(\cdot)$. 
Time integrals are evaluated by trapezoidal quadrature with $\kappa$ samples per segment unless a closed form is available.
SFCs are modeled per segment $s$ as the intersection of halfspaces
\[
\text{SFC}_s \;\triangleq\; \big\{ \mathbf{x}\in\mathbb{R}^3 \;\big|\; \mathbf{n}_{s,h}^\top \mathbf{x} \le b_{s,h}-C_{\text{SFC}},\ \ h=1,\dots,H_s \big\},
\]
where $H_s$ is the number of halfspaces on segment $s$, and $C_{\text{SFC}}=0.2$ is a safety margin.
The total objective is
\[
\begin{aligned}
  J =& ~w_T\!\sum_s T_s + w_{\mathrm{smooth}} \tilde J_{\text{smooth}} + w_{\text{SFC}} J_{\text{SFC}} \\
  &~+ w_v J_v + w_a J_a + w_j J_j,
\end{aligned}
\]
where 
\(
  J_{\text{SFC}} \;=\; \sum_s \int_{0}^{T_s} \sum_{h=1}^{H_s} \phi_\mu\!\big(\,\mathbf{n}_{s,h}^\top \mathbf{x}(t) - b_{s,h} + C_{\text{SFC}}\,\big)\, dt,
\) 
penalizes leaving the SFC, 
$J_v=\sum_s\!\int_0^{T_s}\phi_\mu(\|\mathbf{v}(t)\|^2-v_{\max}^2)\,dt$, 
$J_a=\sum_s\!\int_0^{T_s}\phi_\mu(\|\mathbf{a}(t)\|^2-a_{\max}^2)\,dt$, 
and $J_j=\sum_s\!\int_0^{T_s}\phi_\mu(\|\mathbf{j}(t)\|^2-j_{\max}^2)\,dt$.
Note that the jerk smoothness cost \(\tilde J_{\text{smooth}}\) uses the closed form given in Eq.~\eqref{eq:Jjerk}, which avoids per-sample recomputation.
MIGHTY's weights are $w_T=\num{5e2}$, $w_{\mathrm{smooth}}=\num{e-1}$, $w_{\text{SFC}}=\num{e3}$, $w_v=\num{e3}$, $w_a=\num{e3}$, $w_j=\num{e3}$.
We run 10 trials per method. \textbf{Metrics:}
$R_{\mathrm{succ}}$ [\%] (success rate; collision-free, reaches goal);
$T_{\mathrm{opt}}$ [ms] (local optimization time);
$T_{\mathrm{total}}$ [ms] (total planning time);
$T_{\mathrm{trav}}$ [s] (travel time);
$L_{\mathrm{path}}$ [m] (path length);
$S_{\mathrm{jerk}}=\!\int\!\lVert\mathbf{j}(t)\rVert\,dt$ [m/s$^{2}$] (smoothness);
$\bar{S}_{\mathrm{jerk}}=\sqrt{\tfrac{1}{T}\int\!\lVert\mathbf{j}(t)\rVert^{2}\,dt}$ [m/s$^{3}$] (RMS jerk; time-normalized);
$\rho_{\mathrm{vel}},\,\rho_{\mathrm{acc}},\,\rho_{\mathrm{jerk}}$ [\%] (velocity/acceleration/jerk violations).
We report $\bar{S}_{\mathrm{jerk}}$ to account for differing travel times across methods.

Fig.~\ref{fig:static_env_and_path} illustrates the environment and one MIGHTY trajectory (in red in the bottom plot), and Table~\ref{tab:local_trajectory_optimization_benchmarking} gives the full results.
We evaluate each method with its default sensor; for methods whose default is not LiDAR (EGO-Swarm, EGO-Swarm2), we also run a LiDAR variant to align with SUPER and MIGHTY (both LiDAR-based). 
Note that EGO-Swarm with LiDAR performed substantially worse, so Table~\ref{tab:local_trajectory_optimization_benchmarking} omits those data for clarity.
For EGO-Swarm, $v_{\max}=\SI{4.0}{m/s}$ yielded a \SI{40}{\percent} success rate; reducing $v_{\max}$ to \SIrange{3.0}{1.0}{m/s} improved safety, reaching up to \SI{90}{\percent} at \SI{1.0}{m/s}.
For EGO-Swarm2, with depth-camera input, it achieves \SI{100}{\percent} success and low compute time at both obstacle weights \(w_{\text{obst}}\in\{\num{1e4},\num{5e4}\}\). 
With LiDAR input, the success rates are \SI{20}{\percent} and \SI{70}{\percent} at the same weights.
For SUPER, $w_t=\num{e4}$ gave \SI{20}{\percent} success, and reducing $w_t$ to $\num{e1}$ achieved \SI{100}{\percent}. 
Note that SUPER applies a soft penalty on the distance between the trajectory and points in the overlap of successive SFC segments, with weight $\num{5e6}$, which is the largest among all cost terms.
Summarizing the results, we see that EGO-Swarm2 is the fastest in local optimization and total replanning time (\best{\SI{1.2}{ms}} and \best{\SI{1.7}{ms}}), whereas
MIGHTY achieves the shortest travel time (\best{\SI{79.0}{s}}) and path length (\best{\SI{310.9}{m}}). 
The jerk of MIGHTY is higher ($S_{\mathrm{jerk}}=\SI{522.2}{m/s^2}$, $\bar{S}_{\mathrm{jerk}}=\SI{9.4}{m/s^3}$); however, this is consistent with prior observations of the trade-off between performance and smoothness.

\subsection{Dynamic Environments}\label{subsec:dynamic_env_sim}

\begin{figure}[t]
  \centering
  \includegraphics[width=\columnwidth]{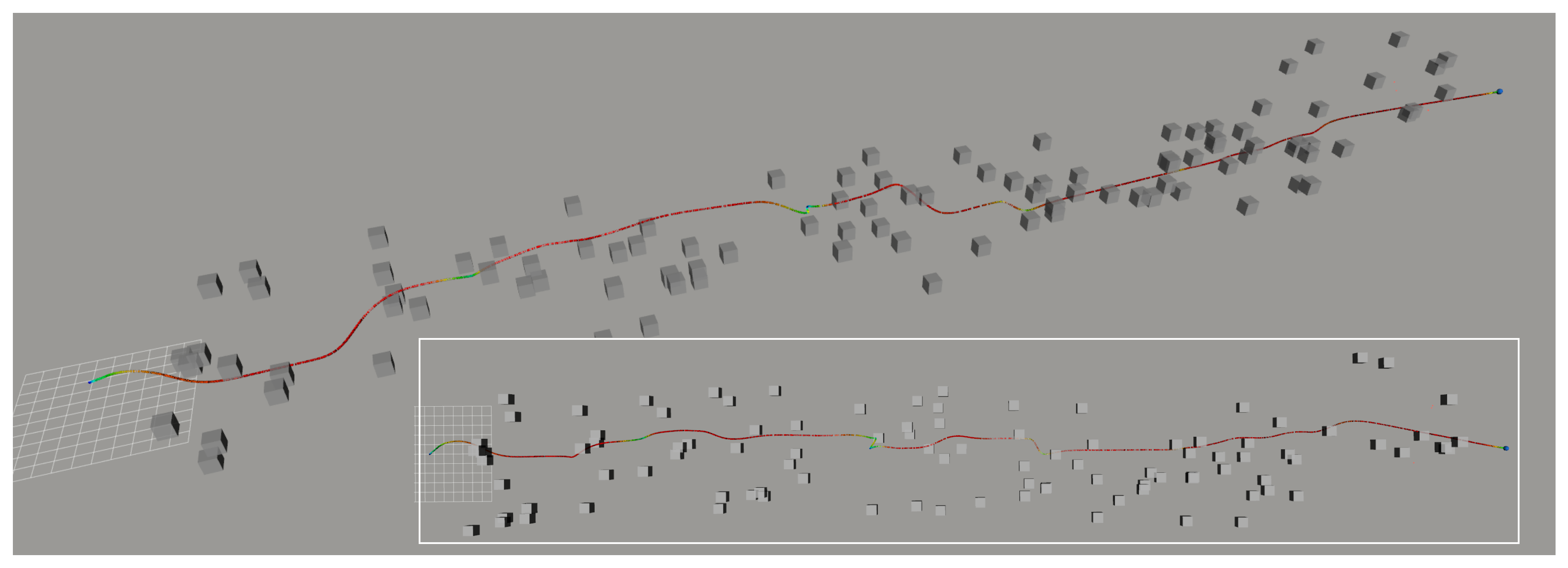}
  \caption{Benchmarking scenario with 100 dynamic obstacles following smooth trefoil-knot trajectories. The image shows the agent's trajectory (color-mapped by speed) navigating through the dynamic obstacles.}
  \label{fig:dynamic_obstacle_only_results}
  \vspace{-0.5em}
\end{figure}

\begin{figure}[t]
  \centering
  \includegraphics[width=\columnwidth]{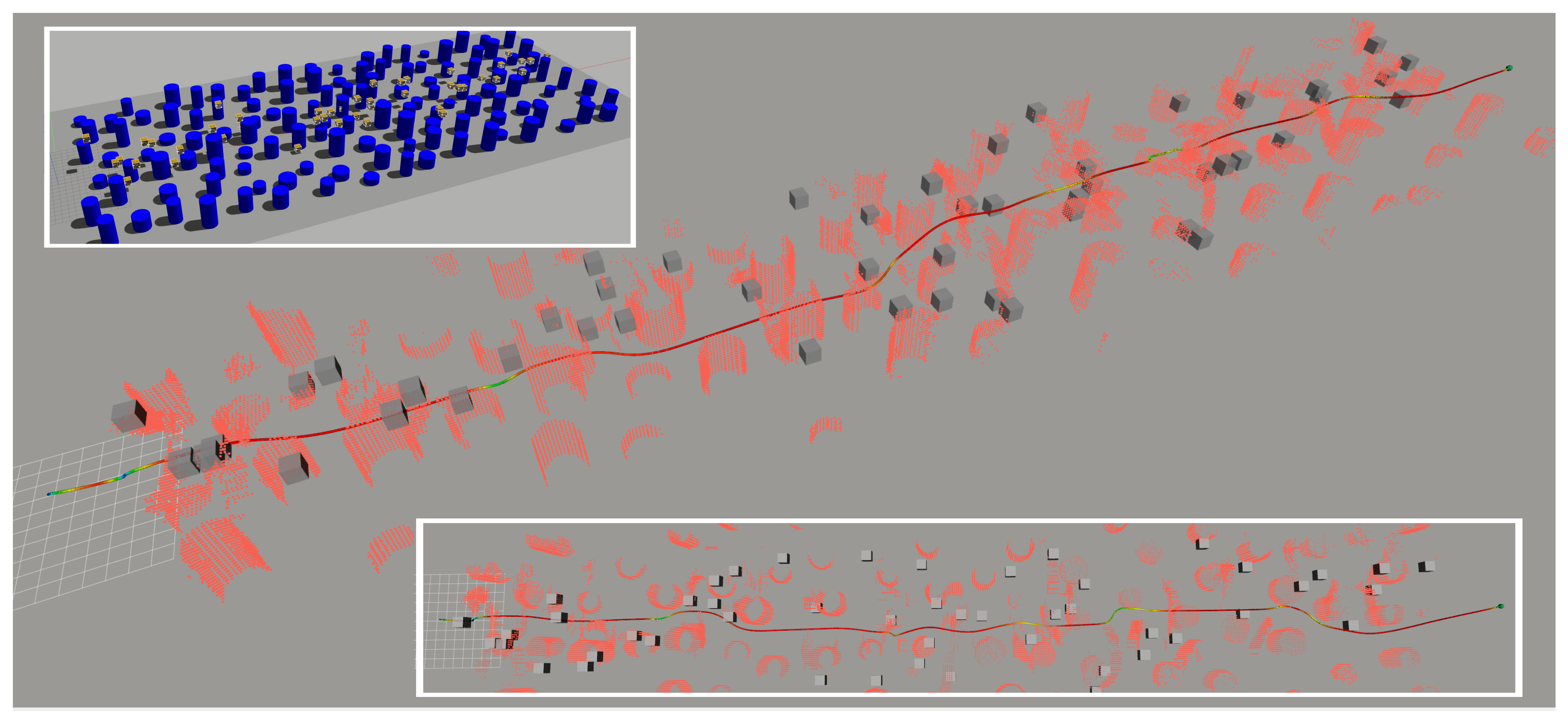}
  \caption{Benchmarking scenario with 50 dynamic obstacles and static obstacles. The top left image illustrates the Gazebo simulation environment. The main image and bottom right image show the point cloud (pink-orange) and agent's trajectory (color-mapped by speed) navigating through both static and dynamic obstacles.}
  \label{fig:dynamic_forest_results}
  \vspace{-0.5em}
\end{figure}

We further evaluate MIGHTY in two dynamic scenarios: (1) only dynamic obstacles, and (2) dynamic \& static obstacles.
For both, in addition to the cost terms described in Sec.~\ref{subsec:benchmarking_static_envs}, we add a dynamic-obstacle avoidance cost
\(
  J_{\text{dyn}} = \int_{0}^{\sum_s T_s} \big(C_{\text{dyn}}^2 - \|\mathbf{x}(t)-\mathbf{k}(t)\|^2\big)_+^{3}\,dt.
\)
where $\mathbf{k}(t)$ is the position of a dynamic obstacle at time $t$, and $(\cdot)_+=\max(0,\cdot)$.
We set the weight $w_{\text{dyn}}=\num{e1}$ and use a soft barrier radius $C_{\text{dyn}}=\SI{3.0}{m}$.
Note that $C_{\text{dyn}}$ is where the penalty activates, not a hard safety distance.
The robot collision radius is \SI{0.1}{m}. 
The start and goal are $(0,0,3)\,\si{m}$ and $(100,0,3)\,\si{m}$, and, in this more complex environment, the dynamic constraints are set to $v_{\max}=\SI{2.0}{m/s}$, $a_{\max}=\SI{5.0}{m/s^2}$, and $j_{\max}=\SI{30.0}{m/s^3}$.

\textbf{Dynamic-only environment:} 
We generate 100 dynamic obstacles following smooth trefoil-knot trajectories, with initial positions $x\in[0,100]\,\si{m}$, $y\in[-10,10]\,\si{m}$, $z\in[1,6]\,\si{m}$. 
Fig.~\ref{fig:dynamic_obstacle_only_results} shows an example of the environment and the trajectory. MIGHTY avoids all obstacles in \best{10}/10 trials. 
The minimum nearest-obstacle distance is \SI{0.8}{m}, above the \SI{0.1}{m} safety threshold.

\textbf{Dynamic \& static environment:} 
We add 50 dynamic obstacles (same motion model) to the static forest as shown in Fig.~\ref{fig:dynamic_forest_results} and run 10 trials. MIGHTY avoids all obstacles in \best{all trials}. 
The minimum is \SI{1.0}{m}, above the \SI{0.1}{m} collision radius.

\section{Hardware Experiments}\label{sec:hardware-experiments}

We also evaluated MIGHTY in three hardware experiments: (1) long-duration flights at \(\,v_{\max}\in\{1,2,3,4\}\,\si{m/s}\), (2) high-speed flights up to \(\,v_{\max}=7\,\si{m/s}\), and (3) flights with obstacles introduced during the mission.
The perception uses a Livox Mid-360 LiDAR, and localization is provided by DLIO~\cite{chen2023dlio}.
Planning runs onboard on an Intel\textsuperscript{\texttrademark} NUC~13, and low-level control uses PX4~\cite{meier2015px4} on a Pixhawk flight controller~\cite{meier2011pixhawk}.
All perception, planning, and control run in real time onboard.

\begin{figure}[t]
    \centering
    \includegraphics[width=\columnwidth]{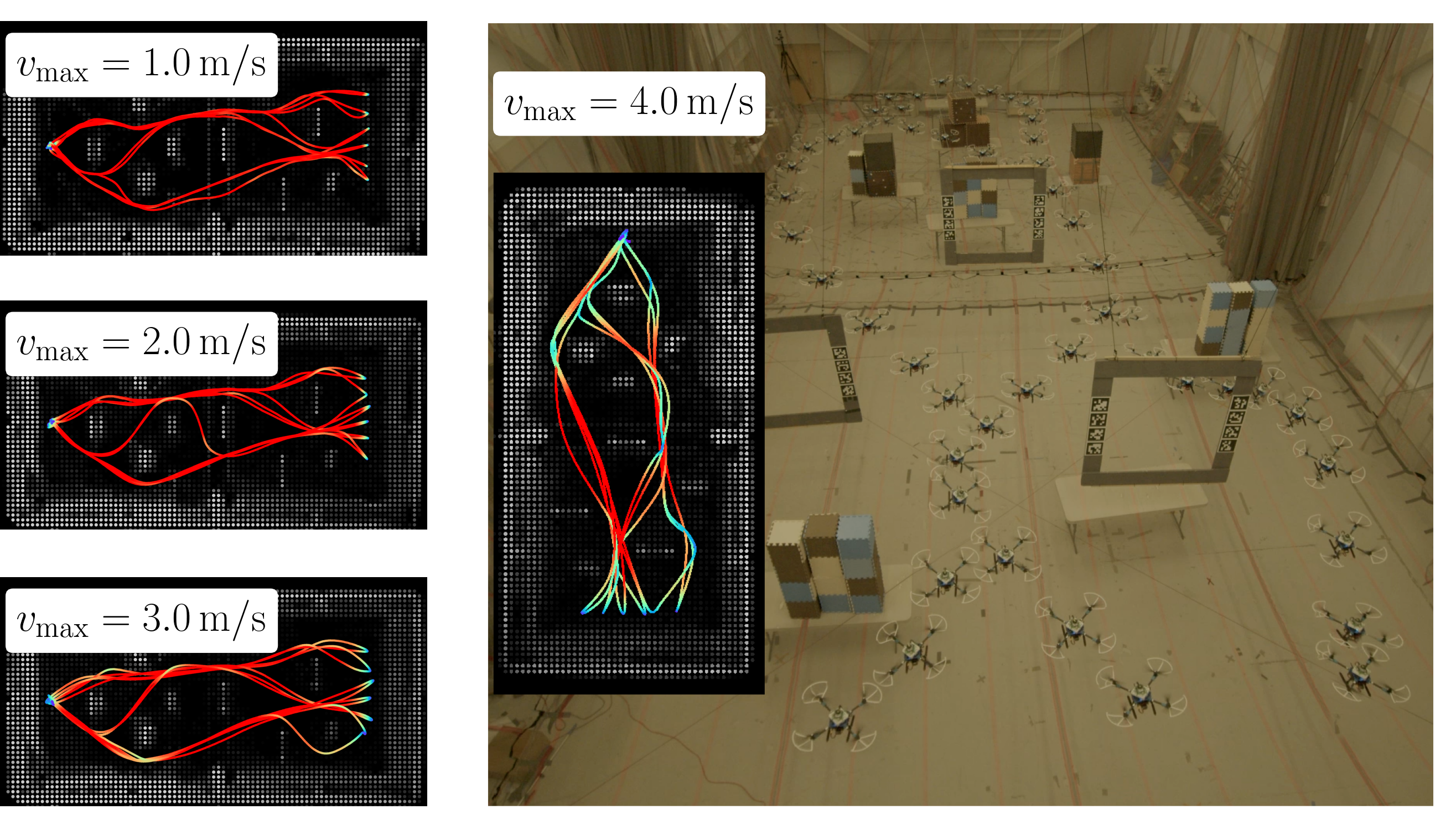}
    \caption{Long-duration experiment with \(v_{\max}\in\{1,2,3,4\}\,\si{m/s}\).
    Top: action sequence at \(v_{\max}=\SI{4.0}{\m/\s}\).
    Bottom: trajectory histories overlaid on the recorded LiDAR point cloud.
    Warmer colors indicate higher speed (red: \(v_{\max}\), blue: \SI{0.0}{\m/\s}); white points are the onboard point cloud.}
    \label{fig:hw_long_duration_flight}
    \vspace{-0.5em}
\end{figure}

\textbf{Long-Duration Flight Experiment}: To assess reliability over extended operations, the vehicle repeatedly traverses a space with multiple obstacles (Fig.~\ref{fig:hw_long_duration_flight}).
Six goal positions are placed opposite the start, inducing repeated out-and-back motions through the environment.
For \(\,v_{\max}\in\{1,2,3,4\}\,\si{m/s}\), all flights completed without collision.

\begin{figure}[t]
    \centering
    \includegraphics[width=\columnwidth]{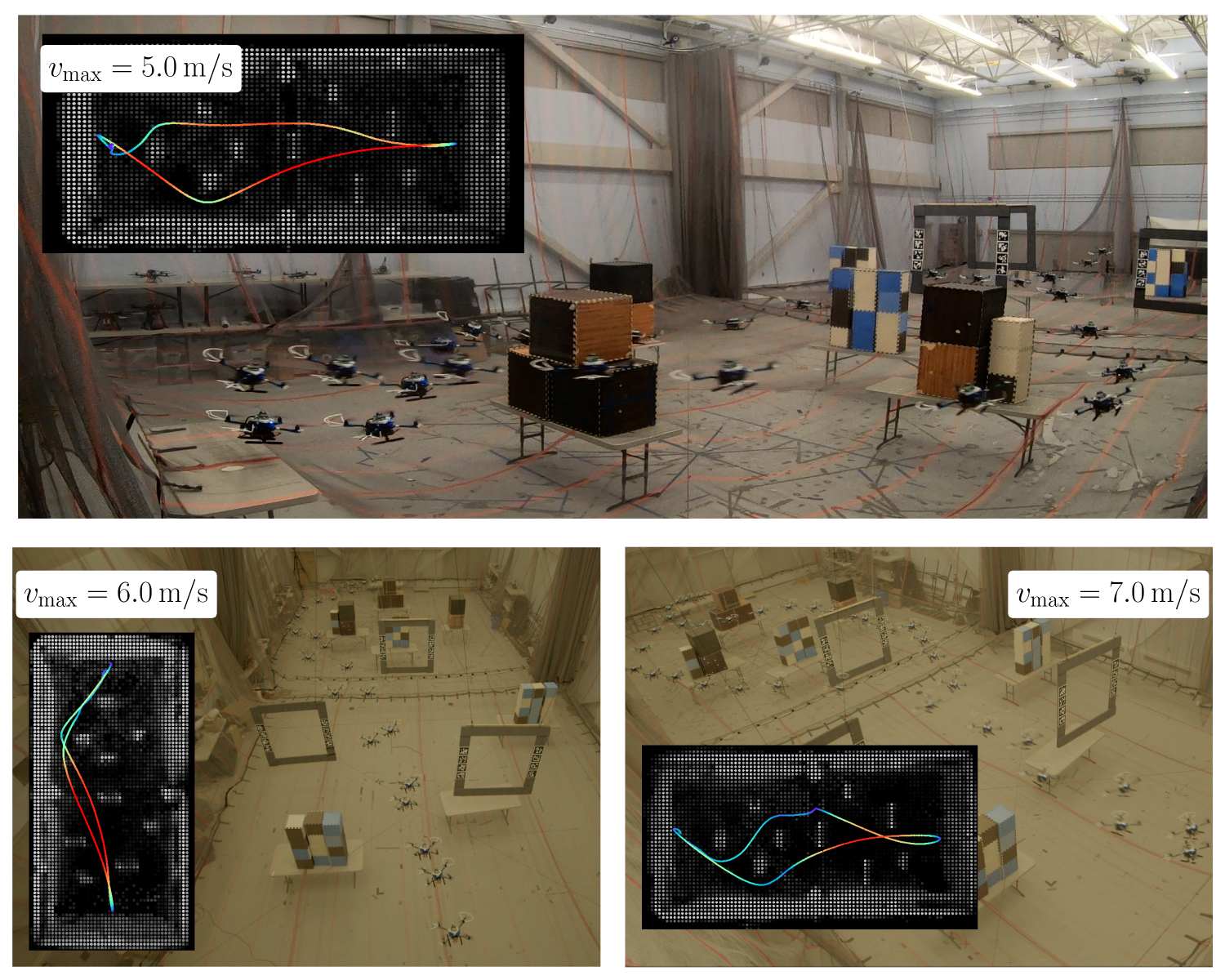}
    \caption{High-speed experiments at \(v_{\max}=\SI{5.0}{}, \SI{6.0}{}, \SI{7.0}{\m/\s}\).
    Each panel shows the action sequence with trajectory history over the recorded point cloud.}
    \label{fig:hw_fast_flight}
    \vspace{-0.7em}
\end{figure}

\textbf{High-Speed Flight Experiment}: We further evaluate performance at higher speeds.
As shown in Sec.~\ref{sec:simulation-results}, MIGHTY tends to produce higher-jerk trajectories; when constraints are satisfied, these trajectories remain safe while enabling faster motion.
We test \(\,v_{\max}=\SI{5.0}{\m/\s}, \SI{6.0}{\m/\s}, \SI{7.0}{\m/\s}\).
Fig.~\ref{fig:hw_fast_flight} shows the resulting histories.
All flights completed without collision, demonstrating that the MIGHTY planner and DLIO state estimation operate effectively at high speed.
At \(\,v_{\max}=\SI{7.0}{\m/\s}\), the vehicle reached \best{\(\SI{6.7}{\m/\s}\)} peak speed.

\begin{figure}[h]
    \centering
    \includegraphics[width=\columnwidth]{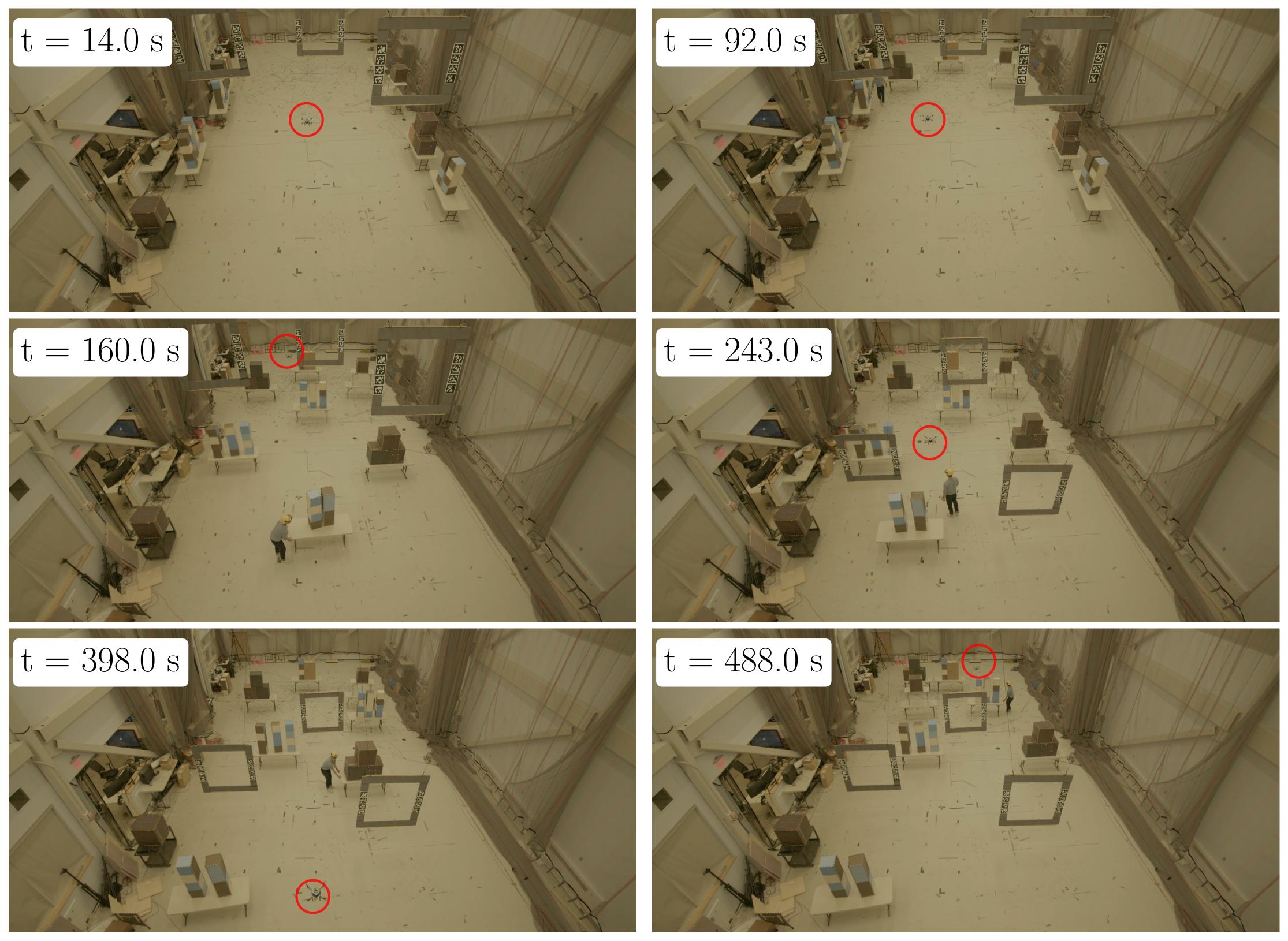}
    \caption{Dynamic-environment experiment at \(v_{\max}=\SI{1.0}{\m/\s}\).
    The red circle shows the UAV; obstacles are introduced and moved during flight.}
    \label{fig:hw_test13}
    \vspace{-0.5em}
\end{figure}

\textbf{Dynamic Obstacle Flight Experiment}: To assess robustness to changes in the environment, we introduce and move obstacles during the mission (Fig.~\ref{fig:hw_test13}).
We track a person carrying obstacles and incorporate the estimates into the trajectory optimization; see Sec.~\ref{subsec:dynamic_env_sim} for details.
The agent successfully reaches the goal without collision over \(\SI{490}{\s}\).

\section{Conclusions}\label{sec:conclusion}
We presented MIGHTY, a Hermite-spline trajectory planner for joint spatiotemporal optimization. 
By optimizing path geometry and time together with local derivative control, MIGHTY generates smooth, collision-free, and dynamically feasible trajectories for high-performance maneuvers. 
Benchmarks and hardware tests show MIGHTY outperforms baselines in travel time while satisfying constraints.

\section{Acknowledgements}\label{sec:acknowledgements}
The authors would like to thank Lili Sun for her help with the hardware experiments and Mason B. Peterson for his support with \textit{robotdatapy} library. 
The authors also thank Juan Rached for his help with hardware setup.
This work was supported in part by DSTA and ARL grant W911NF-21-2-0150.

\bibliographystyle{ieeetr}
\bibliography{root}

\end{document}